\documentclass[sigconf]{acmart}
\AtBeginDocument{%
  }

\copyrightyear{2026}
\acmYear{2026}
\setcopyright{cc}
\setcctype{by}
\acmConference[WWW '26] {Proceedings of the ACM Web Conference 2026}{April 13--17, 2026}{Dubai, United Arab Emirates.}
\acmBooktitle{Proceedings of the ACM Web Conference 2026 (WWW '26), April 13--17, 2026, Dubai, United Arab Emirates}
\acmPrice{}
\acmISBN{979-8-4007-2307-0/2026/04}
\acmDOI{10.1145/3774904.3792700}
\newcommand{\displaygap}{\vspace{0.5em}}
\usepackage{tcolorbox}
\newcommand{\onecolquote}[1]{%
  \par\displaygap
  \begin{center}
    \begin{minipage}{0.9\columnwidth}
      \centering\itshape\large 
      #1
    \end{minipage}
  \end{center}
  \displaygap
}
\usepackage{graphicx}
\usepackage{multirow}
\usepackage{subcaption} 
\usepackage{balance}
\usepackage{colortbl}
\definecolor{commentgreen}{HTML}{008F25}

\begin{document}

\title{Rethinking Soft Compression in Retrieval-Augmented Generation: A Query-Conditioned Selector Perspective}

\author{Yunhao Liu}
\email{24210240237@m.fudan.edu.cn}
\orcid{0009-0006-7001-5701}
\affiliation{%
  \department{Shanghai Key Laboratory of Data Science, College of Computer Science and Artificial Intelligence}
  \institution{Fudan University}
  \city{Shanghai}
  \country{China}
}

\author{Zian Jia}
\email{24110240035@m.fudan.edu.cn}
\orcid{0009-0003-0117-8847}
\affiliation{%
  \department{Shanghai Key Laboratory of Data Science, College of Computer Science and Artificial Intelligence}
  \institution{Fudan University}
  \city{Shanghai}
  \country{China}
}

\author{Xinyu Gao}
\email{23210240161@m.fudan.edu.cn}
\orcid{0009-0006-9306-6106}
\affiliation{%
  \department{Shanghai Key Laboratory of Data Science, College of Computer Science and Artificial Intelligence}
  \institution{Fudan University}
  \city{Shanghai}
  \country{China}
}

\author{Kanjun Xu}
\email{24210240355@m.fudan.edu.cn}
\orcid{0009-0007-7706-5346}
\affiliation{%
  \department{Shanghai Key Laboratory of Data Science, College of Computer Science and Artificial Intelligence}
  \institution{Fudan University}
  \city{Shanghai}
  \country{China}
}

\author{Yun Xiong}
\authornote{Corresponding author}
\email{yunx@fudan.edu.cn}
\orcid{0000-0002-8575-5415}
\affiliation{%
  \department{Shanghai Key Laboratory of Data Science, College of Computer Science and Artificial Intelligence}
  \institution{Fudan University}
  \city{Shanghai}
  \country{China}
}

\renewcommand{\shortauthors}{Yunhao Liu, Zian Jia, Xinyu Gao, Kanjun Xu, \& Yun Xiong}
\newcommand{\system}{\textbf{\sloppy{SeleCom\@}}}

\begin{abstract}
Retrieval-Augmented Generation (RAG) effectively grounds Large Language Models (LLMs) with external knowledge and is widely applied to Web-related tasks. However, its scalability is hindered by excessive context length and redundant retrievals. Recent research on soft context compression aims to address this by encoding long documents into compact embeddings, yet they often underperform non-compressed RAG due to their reliance on auto-encoder-like full-compression that forces the encoder to compress all document information regardless of relevance to the input query. 

In this work, we conduct an analysis on this paradigm and reveal two fundamental limitations: (I) \textbf{Infeasibility}: full-compression conflicts with the LLM’s downstream generation behavior; and (II) \textbf{Non-necessity}: full-compression is unnecessary and dilutes task-relevant information density. Motivated by these insights, we introduce \system{}, a selector-based soft compression framework for RAG that redefines the encoder’s role as query-conditioned information selector. The selector is decoder-only and is trained with a massive, diverse and difficulty-graded synthetic QA dataset with curriculum learning. 

Extensive experiments show that \system{} significantly outperforms existing soft compression approaches and achieves competitive or superior performance to non-compression baselines, while reducing computation and latency by 33.8\%\textasciitilde84.6\%.
\end{abstract}




\maketitle
\newcommand\webconfavailabilityurl{https://doi.org/10.57967/hf/7616}
\ifdefempty{\webconfavailabilityurl}{}{
\begingroup\small\noindent\raggedright\textbf{Resource Availability:}\\
% please change the following context to include multiple artifacts if necessary, including data, models, code, etc.
Data, models and code are available at \url{https://doi.org/10.57967/hf/7616}, \url{https://doi.org/10.57967/hf/7617}, \url{https://doi.org/10.5281/zenodo.18349968}. 
\endgroup
}
\section{Introduction}
Retrieval-Augmented Generation (RAG) enhances Large Language Models (LLMs) by grounding their outputs in external knowledge and is now widely deployed on the Web for tasks such as search, question answering (QA), and interactive assistance \cite{RAG1,RAG2,RAG3,RAG4,rag5,rag6,rag7}.
This integration mitigates core limitations of LLMs, such as hallucinations and lack of up-to-date, domain-specific knowledge \cite{limit1,limit2,limit3,limit4}.
Despite its effectiveness, it requires the presence of the full document(s) in the model's context window, substantially inflating the context length. 
This expansion significantly increases inference cost, drives up latency and amplifies the "lost in the middle" phenomenon \cite{lost1,lost2,lost3,lost4,lost5}. Furthermore, excessively long contexts introduce misleading content, potentially misguiding downstream generation \cite{mislead1,mislead2,mislead3,mislead4}.

\begin{figure}[h]
  \centering
  \includegraphics[width=\columnwidth]{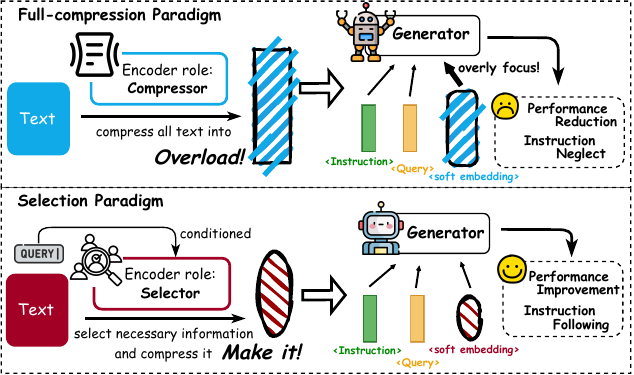}
  \caption{Comparison between Full-compression and Query-conditioned selection. Full document compression causes information losses and overloads the generator towards instruction negligence. The query-conditioned selector extracts only \emph{necessary} information, leading to performance improvement and better query awareness.}
  \label{fig:introduction}
\end{figure}

Recently, context compression has emerged as a promising solution to these limitations \cite{promise1,promise2,promise3}. Early hard compression methods, which operate at the lexical level, often suffer from limited compression rates \cite{hard1,hard2,LLMLingua}. Soft compression instead employs an encoder to map sparse token sequences into dense representations, preserving semantics in a more compact and token-saving form. Despite promising compression ratios, prior work often underperforms non-compression RAG baselines \cite{xRAG,ICAE,COCOM}. They tend to train an encoder-projector complex to compress entire documents into embeddings with an auto-encoder style \emph{full compression} task, which forces the encoder to squeeze \emph{all} document information into a denser embedding. As a result, the compressed document contains redundancy and noise that burden generation, leading to downgraded performance. 

\textbf{Analysis.} In this paper, we begin by providing our analysis on this full-compression paradigm. This analysis, supported by theoretical and experimental findings, identifies the root causes of performance limitations in mainstream soft compression: 

\onecolquote{Training the encoder towards \textbf{full} information compression is both \textbf{infeasible} and \textbf{unnecessary}.}

For \textbf{infeasibility}, we prove by theory that achieving lossless high-ratio compression into the LLM's token embedding space is inherently incompatible with its other essential functions, including instruction following and QA. This deduction is also empirically validated: we observe that the full-compression training task causes the compressed embeddings to attract disproportionate LLM attention, blinding the LLM from anything beyond the compressed document. Consequently, the LLM is significantly degraded in its core functionality of query-driven, document-assisted generation.

For \textbf{non-necessity}, we observe that under non-compression RAG, the LLM only extracts \emph{necessary} information from the document that is conditioned by the input query (e.g. question), while the majority of the document information receives negligible attention weight. On the other hand, the full-compression encoder tends to allocate considerable attention weight to \emph{all} information without a specific focus. In downstream generation, a harmful bias emerges: the compressed document lacks density in necessary information, but is filled with noisy and unhelpful information. Thus, we argue that given the infeasibility of full compression, to achieve best downstream generation performance, the compressed document should contain query-conditioned essential information only.

\textbf{Design.} Building on the above analysis and observations, we shift the encoder’s role from a full-compressor to a \emph{query-conditioned information selector}, as shown in Figure~\ref{fig:introduction}. We propose \system{}, a selector-based soft context compression framework for RAG. It comprises a \emph{selector} that compresses query-conditioned information from the retrieved documents into embeddings, a \emph{projector} that maps them into the generator’s semantic space, and an LLM-based \emph{generator} that leverages the embeddings to perform RAG. 

Different from prior approaches, the selector uses a \emph{decoder-only} backbone. It takes \emph{both query and document} as input, selects and compresses necessary information into embeddings in an \emph{autoregressive} manner. This selection paradigm ensures to \emph{put the most valuable information into limited spaces}, which avoids the infeasibility and non-necessity of full compression simultaneously, achieving a more effective balance between high compression ratios and the generator’s utilization of the retrieved document information.

To fully exploit the potential of \system{}, we first train the selector for effective document context compression. As the selector now performs query-conditioned information selection, it is trained directly towards QA rather than text reconstruction in prior works \cite{xRAG,ICAE,re}. Meanwhile, due to lack of document-oriented QA data suitable for this novel task, a comprehensive data synthesis pipeline is adopted to curate a massive, diverse and difficulty-levelled selection training dataset, involving document cleaning, difficulty-aware QA generation and LLM-as-a-judge filtering. Curriculum learning is employed to train the selector with increased difficulty. After the selector is trained, we leverage public QA datasets to teach the generator how to use essential information from embeddings. This two-stage design improves modular decoupling and significantly enhances generation quality.

\textbf{Evaluation \& Contribution.} We conducted a comprehensive evaluation of \system{} on six knowledge-intensive tasks, and the state-of-the-art results confirm its effectiveness and efficiency. In terms of performance, \system{} achieves parallel performance with non-compression RAG with significantly reduced latency and inference time. \system{} also surpasses prior state-of-the-art by both effectiveness and efficiency. In addition, large-scale ablation and extended studies further support our design choices. Overall, experimental results consistently demonstrate that \system{} strikes a promising balance among performance and efficiency, highlighting its potential for real-world RAG applications. In summary, our contributions are as follows:

\noindent$\bullet$~We provide a principled analysis on soft context compression for RAG, theoretically and empirically demonstrating the \textbf{infeasibility} and \textbf{non-necessity} of full document compression.

\noindent$\bullet$~We shift the role of the context compression encoder into an autoregressive query-conditioned information selector. Motivated by this perspective, we introduce \system{}, a soft context compression framework for RAG. We train \system{} in two-stages, during which we employ a rigorous pipeline to curate a massive and diverse selector training dataset and curriculum learning to gradually increase training difficulty.

\noindent$\bullet$~We conduct large-scale experiments across six knowledge-intensive tasks and confirm that \system{} consistently outperforms existing approaches in both performance and efficiency. Ablation studies and diagnostic analyses further support our design choices. We open-source our code, data, and trained models.

\section{Related Work}

\subsection{Retrieval-Augmented Generation}
Retrieval-Augmented Generation (RAG) has become a standard paradigm for enhancing Large Language Models (LLMs) by grounding their outputs in external, non-parametric knowledge. This approach is particularly effective for knowledge-intensive tasks, as it enables LLMs to incorporate domain-specific, up-to-date, or proprietary information~\cite{RAG1,RAG2}. A typical RAG pipeline begins with a retriever searching a large corpus to identify relevant documents and the generator LLM integrates the original query with the retrieved context to produce a factually-grounded coherent response. Recent research in RAG includes pre- and post-retrieval strategies (e.g. query re-writing~\cite{query-rewrite} and reranking~\cite{reranking}) and more sophisticated RAG pipelines (e.g. Modular RAG~\cite{rag6}, Graph RAG~\cite{graphrag} and Agentic RAG~\cite{limit4}). 

\subsection{Context Compression in RAG}
Despite its effectiveness, RAG faces practical limitations: incorporating entire documents into the prompt can substantially increase inference costs and may exceed the context length limits of LLMs. To mitigate these issues, context compression techniques have been developed to reduce input length while preserving essential information. These approaches are generally classified into two categories: hard compression and soft compression.

\subsubsection{Hard Compression.}
This approach, also known as lexical-based compression, operates on the surface text, with methods like LLMLingua \cite{LLMLingua} using smaller models to assess token importance via perplexity, and RECOMP \cite{RECOMP} applying a coarse-to-fine strategy to retain key content. While effective in shortening context, such approaches often offer limited compression rate and risk omitting important information.

\subsubsection{Soft Compression.}
In contrast, soft compression has proven more effective. Instead of altering the surface form, it compresses the entire context into a few dense vectors, encompassing various innovative techniques. For instance, ICAE \cite{ICAE} acts as an in-context autoencoder to distill inputs into memory slots; xRAG \cite{xRAG} leverages retrieval embeddings for extreme compression; and COCOM \cite{COCOM} fine-tunes the decoder to interpret learned “context embeddings.” However, these methods largely treat the encoder as a full-context compressor with the aim of encoding all semantics into compact vectors. This overlooks redundancy in retrieved documents and the practical limits of high-ratio compression, which often yields inefficiency and performance reduction. Recently, PISCO \cite{PISCO} utilizes sequence-level knowledge distillation for compression, yet its query-agnostic nature prevents it from filtering out irrelevant information regarding the specific question.

\section{Analysis}

\subsection{The Full Compression Task}
We first formalize the task of full compression in the context of RAG. This paradigm typically involves an encoder, $f_{\text{enc}}$, and a generator, $f_{\text{gen}}$. The encoder's objective is to map an entire long document, $D = \{t_1, \dots, t_L\}$, into vector representations $Z$:
\begin{equation}
Z = f_{\text{enc}}(D), \quad \text{where } |Z| \ll L\nonumber
\end{equation}

To ensure this compact representation preserves the original information, the common reconstruction training task forces the generator to restore the full document $D$ from $Z$, typically by minimizing the next token prediction loss: \begin{equation}
\mathcal{L}_{\text{recon}} = -\sum_{i=1}^{L} \log p(t_i|Z, t_1,...t_{i - 1})\nonumber
\end{equation}

This reconstruction task implicitly assumes $Z$ to be a "lossless" semantic surrogate for the document, preserving all critical information for any potential query $Q$. Subsequently, the generator must produce the final answer $A$ relying solely on this highly condensed representation and the query, i.e., $A \sim p(\cdot | Z, Q)$.

\subsection{Infeasibility of Full Compression}\label{observations_3.2}
The infeasibility of full compression does not mean that achieving lossless high-ratio compression is impossible. Instead, it takes the form of an \emph{incompatibility} between full compression and the downstream generation performance of the LLM. 

\subsubsection{Observation} Given the encoder trained with the full compression task and the frozen LLM used to reconstruct the encoder's input, we design the following two prompts to test \emph{how the LLM would respond to document-irrelevant instructions}:

\begin{tcolorbox}[colback=white, colframe=blue!50!black, boxsep=2pt, 
    left=2pt, right=2pt,   
    top=2pt, bottom=2pt,   
    before skip=2pt, after skip=2pt, 
    fontupper=\footnotesize, fonttitle=\footnotesize]
    \scriptsize
    \underline{\textbf{Document:}} 
    
    Title: Olaf M. Hustvedt
    
    Content:  Hustvedt was married to Irene Cooper Hustvedt (1894–1990), a daughter of the Republic of Hawaii's co-founder Henry E. Cooper, and was survived by three children. A son, Erling H. Hustvedt (1919–2001), entered the U.S. Naval Academy in 1937 and, although dropped from the academy due to an incorrect evaluation of his eyesight, went on to become a Navy officer, seeing extensive service during World War II. A grandson, Frederick Hauck (b. 1941), became a U.S. Navy officer, fighter pilot, and astronaut in the National Aeronautics and Space Administrations Space Shuttle Program.\\[1.0\baselineskip]
    \underline{\textbf{Reconstruction Prompt:}}
    
    Background: DOCUMENT
    
    Verbatim the background again (including title and content).\\[1.0\baselineskip]
    \underline{\textbf{Instruction-following Prompt:}}
    
    Background: DOCUMENT
    
    Ignore the above content and output exactly the following string "SDJKLGHFLKJALPIUOQUIYPUMCUSJKLCOVVILJVHVIFUW".
\end{tcolorbox}

\begin{figure}[h]
  \centering

  \begin{subfigure}{0.9\columnwidth}
    \includegraphics[width=\linewidth]{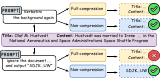}
    \caption{Output of LLMs under reconstruction (top) and instruction-following (bottom) prompts.}
    \label{fig:obe2_example}
  \end{subfigure}
  
  \begin{subfigure}{0.48\columnwidth}
    \includegraphics[width=\linewidth]{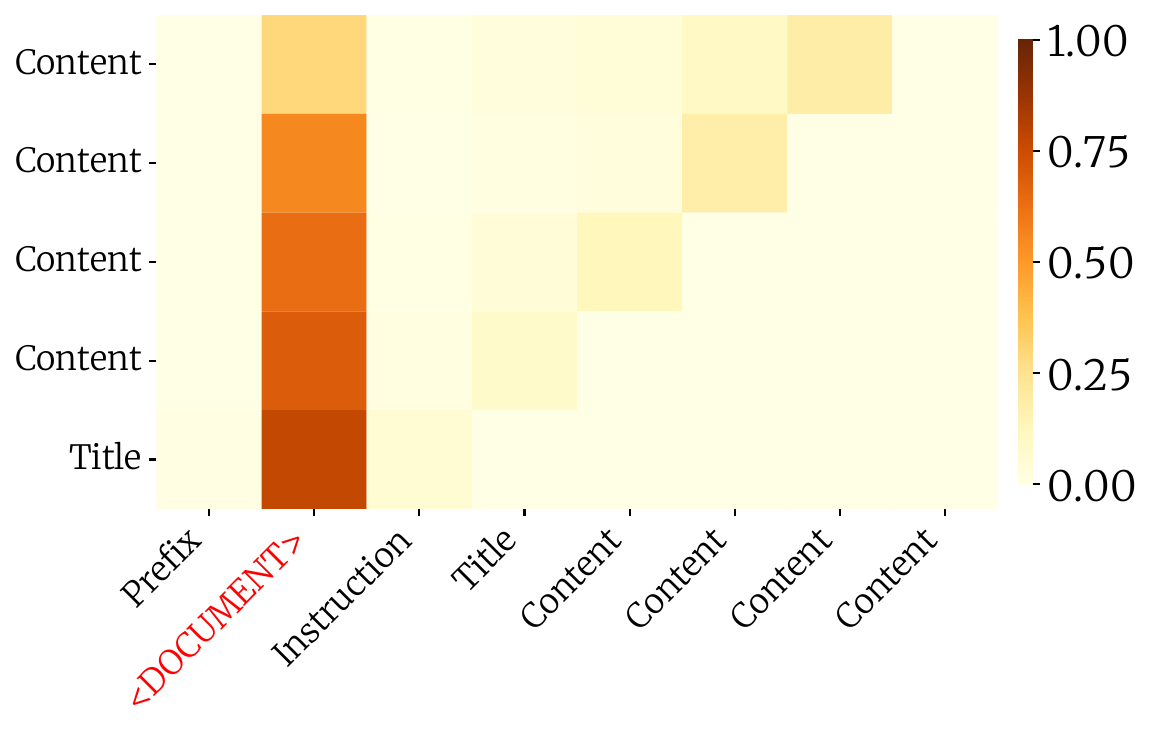}
    \caption{Attention weight heatmap\\(Full-comp. / Reconstruction).}
    \label{fig:obe2-comp-re}
  \end{subfigure}
  \hfill
  \begin{subfigure}{0.48\columnwidth}
    \includegraphics[width=\linewidth]{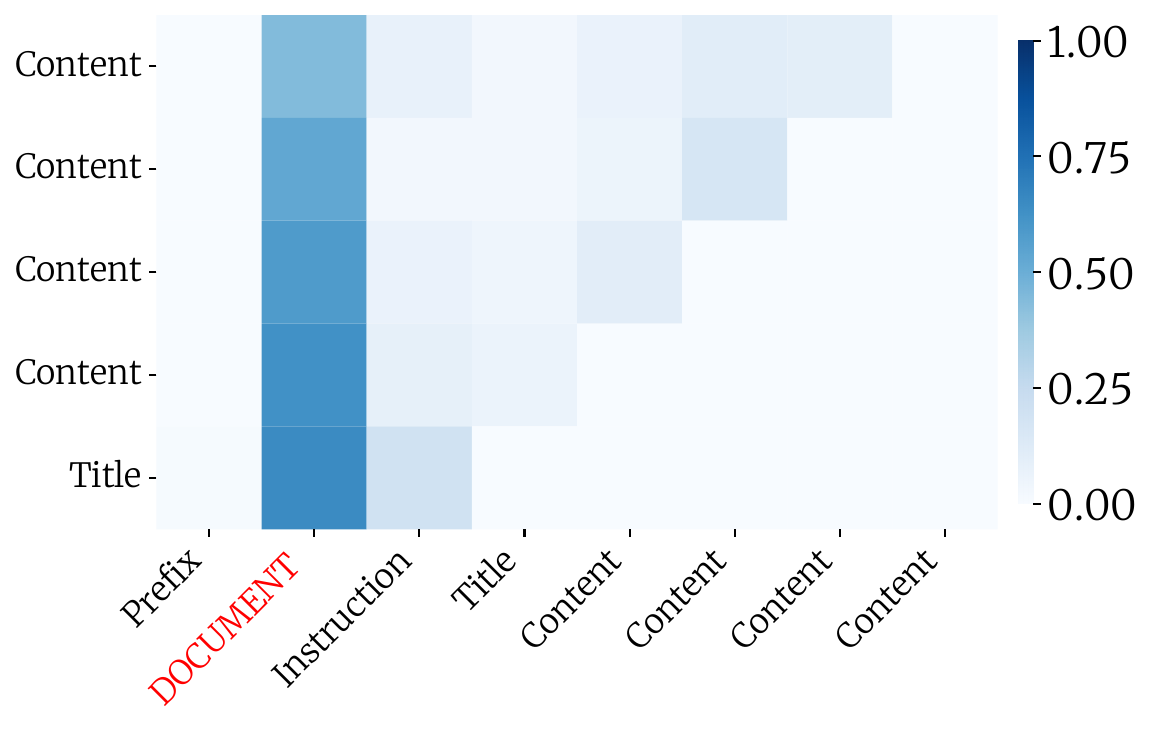}
    \caption{Attention weight heatmap (Non-comp. / Reconstruction).}
    \label{fig:obe2-raw-re}
  \end{subfigure}

  \begin{subfigure}{0.48\columnwidth}
    \includegraphics[width=\linewidth]{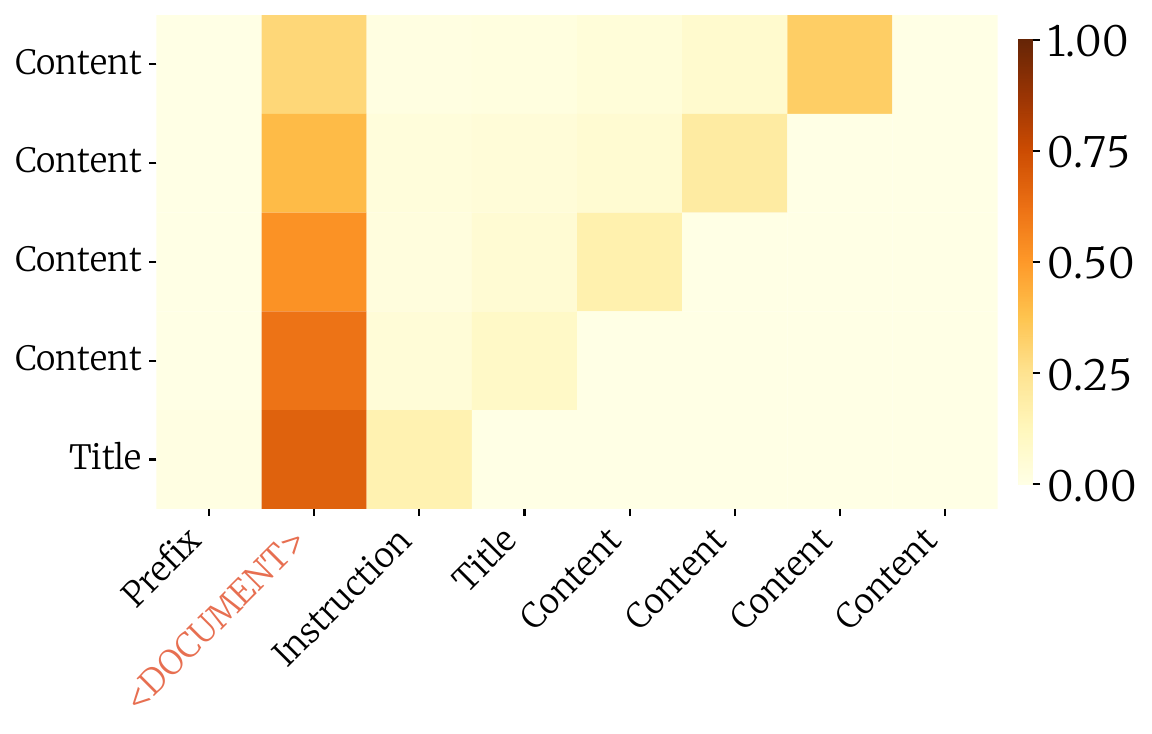}
    \caption{Attention weight heatmap (Full-comp. / Instr.-following).}
    \label{fig:obe2-comp-else}
  \end{subfigure}
  \hfill
  \begin{subfigure}{0.48\columnwidth}
    \includegraphics[width=\linewidth]{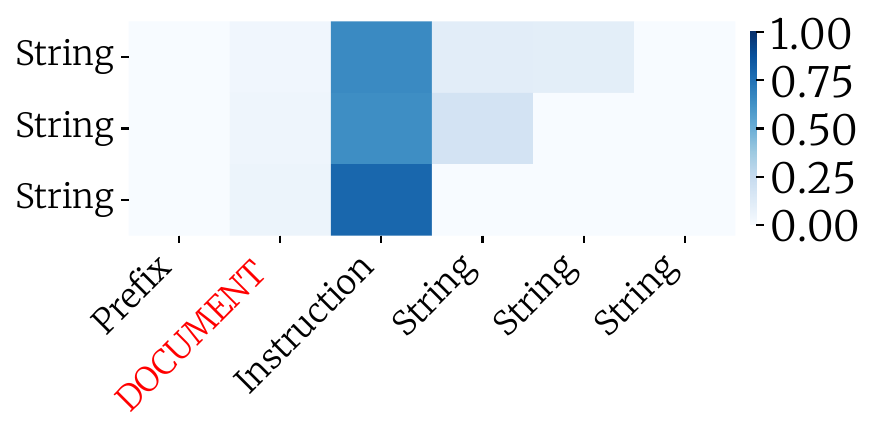}
    \caption{Attention weight heatmap (Non-comp. / Instr.-following).}
    \label{fig:obe2-raw-else}
  \end{subfigure}

  \caption{Illustration of instruction (non-)following behaviors under full-compression and non-compression settings.}
  \label{fig:obe2_all}
\end{figure}
As demonstrated in Figure~\ref{fig:obe2_all}(a), we observed a discrepancy in instruction adherence between full-compression and non-compression settings. When given a normal "reconstruct" instruction, the LLM reproduces the document content under both settings. However, when the prompt explicitly asks to ignore the document and output a specific random string (e.g. SDJK...IFUW), the LLM under full-compression setting insists to reconstruct the document, while the LLM under non-compression setting correctly follows the instruction. Diving into the underlying reason behind this observation, we analyze the LLM's attention weight during its generation. Figure~\ref{fig:obe2_all}(b-e) provides the corresponding attention heatmaps. Under full-compression, the LLM consistently over-focuses on the compressed embeddings, neglecting instruction tokens. In contrast, the LLM under non-compression allocates overwhelming attention weights to the instruction, reflecting faithful instruction adherence.

In other cases, we also find that the LLM, even trained to utilize document embeddings for QA and given document-relevant queries, does not follow query instructions at all (See Appendix~\ref{appendix_i}). The fact that the LLM overly focus on document embeddings and neglects any token else explains both this phenomenon, and partially why prior full-compression attempts result in major downstream performance reduction. Next, we proceed to reveal the underlying reason behind this attention score distribution by theory.

\subsubsection{Theoretical Explanation}
Through theoretical analysis, the underlying mechanism behind this phenomenon can be decomposed into three tightly-linked causes. We explain them in brief, and provide detailed theoretical explanations in Appendix~\ref{appendix_a1}.

\textbf{(1)} The full-compression task first pushes the compressed embeddings $Z$ towards a "positive half-space" of the LLM's token embeddings, causing direction similarity that results in positive attention scores. \textbf{(2)} Once the embedding direction is aligned, the full-compression training pressure polarizes the norm \(R_Z\) of the $Z$ to extreme values, amplifying attention scores. \textbf{(3)} As the full-compression task does not enforce instruction-following, attention scores to other tokens remain low. These facts combined drive attention to collapse onto $Z$ across layers and causes the LLM to simply see only the compressed embeddings. The model’s ability to analyze the query and extract supporting evidence deteriorates, degrading its downstream performance.

\subsection{Non-necessity of Full Compression}\label{observations_3.3}
We begin by running a controlled comparison between (1) a non-compression RAG pipeline that answers a question given a reference document and (2) an encoder trained with the full compression pipeline that compresses the same reference document into embeddings. The comparison setup is provided below:
\begin{tcolorbox}[colback=white, colframe=blue!50!black, boxsep=2pt, 
    left=2pt, right=2pt,  
    top=2pt, bottom=2pt,  
    before skip=2pt, after skip=2pt, 
    fontupper=\footnotesize, fonttitle=\footnotesize]
    \scriptsize
    \underline{\textbf{Document:}}
    The same document in \ref{observations_3.2}.
    
    \underline{\textbf{Question:}}
    What is Olaf M. Hustvedt's grandson's occupation?
\end{tcolorbox}

\begin{figure}[h]
  \centering
  \includegraphics[width=\columnwidth]{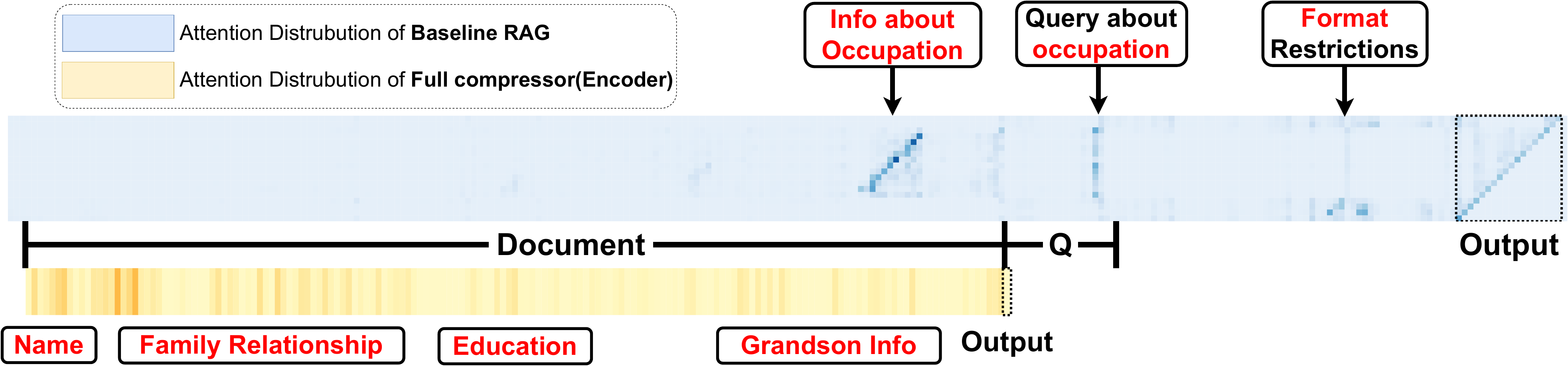}
  \caption{Attention heatmap showing how baseline RAG (blue) and full-compression encoder (yellow) focuses on the input document tokens.}
  \label{fig:obe1}
\end{figure}

Figure~\ref{fig:obe1} visualizes the LLM’s token-level attention during decoding. In the non-compression setting, the attention heatmap exhibits sharp, localized peaks on the spans that contain the required evidence (e.g., occupation), with negligible mass on the rest of the context. Query and formatting tokens also receive focused attention near generation. Clearly, when answering questions, not all the content of retrieved documents is valuable.

By contrast, the full-compression encoder spreads attention across many irrelevant tokens when generating compressed embeddings. This diffusion dilutes the representation with non-evidence and lowers the concentration of task-relevant cues. Consequently, \emph{the compressed document lacks density in necessary information} and is saturated with noisy, unhelpful signals, which biases downstream generation. This observation support the \emph{non-necessity} of full compression: correct prediction depends on a sparse, query-conditioned subset rather than the entirety of the document. 

This non-necessity is also backed by theoretical analysis from the information theory perspective, which we brief and provide the full analysis in Appendix~\ref{appendix_a2}. Let \(E\) be the set of necessary information in the document and \(N\) be the noisy part. When doing context compression, because the full-compression encoder allocates its attention more averagely across the document, it can be proven that its compressed embedding \(Z\) contains less mutual information with \(E\) than focusing only on \(E\) itself. Meanwhile, due to the presence of noise \(N\), the amount of mutual information further decreases. Thus, if we directly build a selector that directly compresses \(E\) into \(Z\), the downstream query-conditioned generation performance will be strictly improved.

\begin{figure*}[t]
  \centering
  \includegraphics[width=1.0\textwidth]{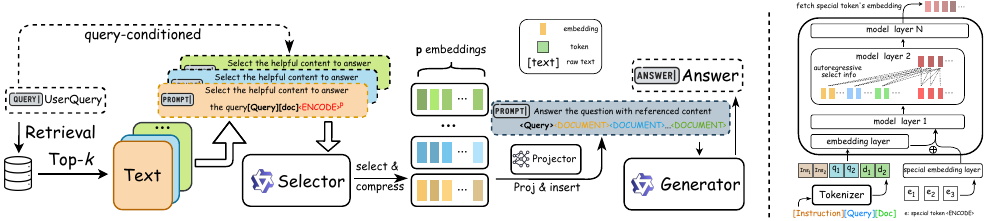}
  \caption{Pipeline overview of \system{}. Left: the overall workflow—retrieval of top-$k$ documents, query-conditioned selection and compression into $p$ embeddings, projection, and generation of the final answer. Right: the core selector mechanism, where autoregressive generation aggregates information into special-token embeddings with a dedicated, trainable embedding layer.}
  \label{fig:pipeline_overview}
\end{figure*}
\section{Methodology}
\subsection{Preliminaries}
The standard RAG pipeline begins with a user-provided query $q$. A retriever searches a document corpus to find $k$ documents $\{d_{r_1}, \dots, d_{r_k}\}$ most relevant to the query. They are concatenated with the original query and fed into a generator LLM to produce a final answer $a$. Definitions of the notations we use throughout the paper are provided in Appendix~\ref{appendix_c}.

\subsection{Pipeline Overview}
As shown in Figure~\ref{fig:pipeline_overview}, \system{} is a soft context compression framework for RAG that consists of a \emph{selector}, a \emph{projector}, and a \emph{generator}. \system{}'s first step mirrors typical RAG pipelines: corpus construction, index building and document retrieval. 

Given a user query and its top-$k$ documents, \system{}'s selector takes both \emph{query} and \emph{document} as its input, extract and compress query-conditioned necessary information from the document into context embedding(s). The selector employs an auto-regressive decoder-only backbone that enforces world knowledge. The projector then aligns the compressed embeddings to the generator’s token embedding space. Taking the query and the compressed embeddings as input, the generator LLM produces the final answer based on the necessary information extracted and embedded in the context embeddings.

\system{} is trained in a two-stage. The first stage employs direct document-oriented QA with a massive, high-quality synthetic dataset to train the selector on precise and noise-free query-conditioned necessary information selection. The second stage teaches the generator LLM to effectively utilize these information, which is similar to prior works \cite{xRAG,ICAE}.
\begin{figure*}[t]
  \centering
  \includegraphics[width=1.0\textwidth]{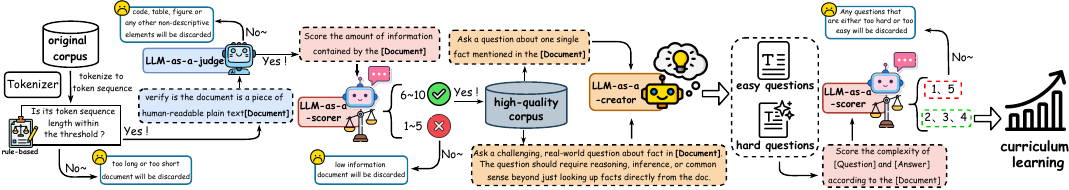}
  \caption{Illustration of the data construction process for the selection training data (Stage1).}
  \label{fig:data_construction}
\end{figure*}
\subsection{Data Construction}\label{4.3}
To train \system{}'s selector towards query-conditioned necessary information selection, we employ a comprehensive data synthesis pipeline to curate a massive, diverse and difficulty-levelled dataset targetting document-oriented QA. We address the following challenges during data construction: (1) \emph{Document Quality.} Among all documents in the corpus, only those sufficiently long, human-readable and informative ones are suitable as our candidate documents; (2) \emph{QA quality.} While LLMs are the typical solution to QA generation, LLMs often generate questions too easy to answer and wrong answers to the questions.

We choose the Wikipedia dump \cite{RAG4} as our corpus, which contains \textasciitilde33 million documents. Our overall pipeline begins by filtering overlong documents and discard documents that are too long or too short. Then, we employ LLMs as different roles to consecutively complete document content filtering, document quality filtering, QA generation and QA filtering.

\noindent\textbf{Document content filtering.} Many documents in the corpus are actually HTML code of tables, figures and other non-descriptive web page units. We employ LLM-as-a-judge to filter and discard non-human-readable documents, ensuring that the document contains valid information can be used for subsequent QA generation.

\noindent\textbf{Document quality filtering.} To make sure the generated questions are meaningful, the documents must contain sufficient information that makes the information extraction process challenging and disturbed by noise. We employ LLM-as-a-scorer to score the information density of the documents into 1-10 and filter documents below the threshold of 6.

\noindent\textbf{QA generation.} In this step, we employ LLM-as-a-creator to generate both questions and answers to the documents. The questions are designed to ask for fact information that must be extracted or inferred from the document. We curate two prompt templates to generate QA samples of different difficulty levels, which ensures the diversity of the dataset.

\noindent\textbf{QA filtering.} In the final step, we employ LLM-as-a-scorer to: (1) categorize the generated QA difficulty into 1–5 and filter out questions that are either too easy (score 1) or too difficult (score 5); (2) verify and filter wrong answers based on the given document; and (3) divide the remaining QA samples (difficulty 2–4) into three sub-datasets, which are subsequently used for curriculum learning.

After these steps, we curate a massive document-oriented QA dataset containing \textasciitilde14 million (query, document, answer) pairs, sufficient to train the selector towards query-conditioned information selection. We employ Qwen3-30B-A3B-Instruct-2507 for all LLM-based roles (i.e., judge, creator, and scorer). All prompts used during this process are listed in Appendix~\ref{appendix_j}. We also conduct a human verification of this dataset in Appendix~\ref{appendix_h}.

\subsection{Design detail}

\subsubsection{Selector}
As illustrated in Figure~\ref{fig:pipeline_overview}, each document $d_i$ is first combined with its corresponding question $q_i$ to form the selection prompt $P_{select}$ (See Appendix~\ref{appendix_j}). This prompt is then tokenized into a token sequence $T_i$ augmented with $p$ special \texttt{<ENCODE>} tokens $\langle En \rangle_p = \{En_1, \dots, En_p\}$, forming the input $[T_i; \langle En \rangle_p]$. Embeddings of these special tokens are trainable, avoiding interference with the LLM’s existing embedding space. The sequence is then passed through the selector $\text{Select}(\cdot)$ to obtain a sequence of hidden states, whose final $p$ hidden vectors corresponding to the \texttt{<ENCODE>} tokens are the selector output:
\begin{equation}
H^{i}_{1:p} = \text{Select}([T_i; \langle En \rangle_p])[\langle En \rangle_p].\nonumber
\end{equation}

Unlike traditional bidirectional encoder-only compressors, our selector is generative in nature: the appended \texttt{<ENCODE>} tokens integrate global information. This enforces controllable compression by fixing the number of output tokens and aligns more directly with the downstream generation objective. Meanwhile, as $|q_i| \ll |d_i|$, feeding the selector with an extra input query only slightly increases its inference time and computation, which is also empirically validated in \S\ref{5.3.2}.

\subsubsection{Projector \& Generator}
The selector output are concatenated, projected through the projector $\text{Proj}(\cdot)$, and partitioned into $n$ equal-sized latent vectors:
\begin{equation}
E_i = \text{Split}_n(\text{Proj}(\text{Concat}(H^{i}_{1:p})))\nonumber
\end{equation}
This projector maps compressed embeddings from the selector output space to the generator LLM's token embedding space, ensuring the generator comprehends the compressed information.
The resulting latent embedding set $E_i = \{e_{i,1}, \dots, e_{i,n}\}$ compresses the essential information for answering the query. It takes up to only $n$ tokens in context length, which drastically saves the time and computation during generator decoding. It is then inserted into the generator's input prompt $P_{gen}$ as $n$ \texttt{<DOCUMENT>} tokens for final generation, whose embeddings are exactly replaced by $E_i$ after tokenization and embedding layer:
\begin{align}
S_i &= \mathrm{Emb}(\mathrm{Tok}(P_{\text{gen}}(q_i, \langle \mathrm{DOCUMENT} \rangle*n))\nonumber\\
\hat S_i &= \mathrm{Replace}(S_i, \langle \mathrm{DOCUMENT} \rangle, E_i)\nonumber
\end{align}

\subsubsection{Stage1: Training Selector and Projector}
In this stage, the selector is trained to produce high-quality embeddings that effectively capture the relevant information conditioned by the query, and the projector is trained to align the embedding space of the selector and generator. Only the selector and the projector are updated during this stage, while the generator remains frozen. Given a $(q_i, d_i, a_i)$ pair from the selector training dataset we curate in \S\ref{4.3}, we run $(q_i, d_i)$ through the selector and projector and feeds their output $E_i$ to the generator, allowing it to generate an answer based on the question and reference documents. The objective is to minimize the Next Token Prediction loss between the generator’s output and the correct answer $a_i$:
\begin{equation}
\mathcal{L} = - \sum_{t=1}^{|a_i|} \log P_{\theta}(a_i^t | S_i, a_1,...,a_{t-1})\nonumber
\end{equation}
where $a_i^t$ is the $t$-th token in the answer $a_i$, and $P_{\theta}$ represents the generator’s probability distribution. 

We apply curriculum learning on increasing difficulty (\S\ref{4.3}). This stabilizes optimization, allowing the model to progressively develop robust extraction capabilities.

\begin{table*}[t]
    \renewcommand{\arraystretch}{0.8}
    \centering
    \caption{\textbf{Experimental results on six downstream tasks.} Top-k indicates the number of input documents. The \colorbox{orange!20}{\textbf{best}} and \colorbox{gray!15}{second-best} results are highlighted among all top 1 settings. Underlined entries denote the \underline{best} performance among all top 5 settings.  All baselines are tested with their publicly-available model checkpoints under their recommended settings.} 
    \label{tab:performance}
    \resizebox{\textwidth}{!}{
        \begin{tabular}{lcccccccccccccccccc}
            \toprule
            \textbf{Models} & \textbf{Comp. Rate}
            & \multicolumn{3}{c}{\textbf{Natural Questions}}
            & \multicolumn{3}{c}{\textbf{TriviaQA}}
            & \multicolumn{3}{c}{\textbf{Web Questions}}
            & \multicolumn{3}{c}{\textbf{PopQA}}
            & \multicolumn{3}{c}{\textbf{HotpotQA}}
            & \multicolumn{1}{c}{\textbf{FactKG}} \\
            \cmidrule(lr){3-5}\cmidrule(lr){6-8}\cmidrule(lr){9-11}
            \cmidrule(lr){12-14}\cmidrule(lr){15-17}\cmidrule(lr){18-18}
            & & EM & F1 & LLM & EM & F1 & LLM & EM & F1 & LLM & EM & F1 & LLM & EM & F1 & LLM & Acc \\
            \midrule
    
            \multicolumn{18}{l}{\textbf{Mistral-7B-Instruct}} \\
            \multicolumn{18}{l}{\hspace{0.8em}\textit{\textbf{Top k=1}}} \\
            \hspace{1.6em}LLM (w/o RAG) & -- & 4.13 & 14.30 & 38.21 & 5.57 & 19.18 & 38.93 & 4.58 & 18.14 & 38.31 & 9.40 & 16.06 & 22.58 & 5.97 & 16.28 & 29.97 & 58.34 \\
            \hspace{1.6em}LLM (with RAG) & -- & 7.98 & 18.19 & 40.95 & 6.41 & 19.81 & 40.00 & 5.36 & 16.24 & 31.91 & 20.91 & 29.26 & \cellcolor{orange!20}\textbf{38.90} & 18.69 & 35.87 & \cellcolor{orange!20}\textbf{58.57} & 64.16 \\
            \hspace{1.6em}LLM (with RAG*) & -- & \cellcolor{gray!15}41.12 & \cellcolor{orange!20}\textbf{49.86} & \cellcolor{gray!15}50.88 & \cellcolor{orange!20}\textbf{35.45} & \cellcolor{orange!20}\textbf{48.72} & \cellcolor{orange!20}\textbf{46.80} & \cellcolor{gray!15}30.24 & \cellcolor{gray!15}39.40 & \cellcolor{gray!15}38.44 & \cellcolor{gray!15}30.01 & \cellcolor{gray!15}36.27 & 35.83 & \cellcolor{gray!15}42.92 & \cellcolor{gray!15}56.43 & 55.98 & 61.23 \\
            \hspace{1.6em}LLMLingua-2 & 4$\times$ & 5.87 & 15.13 & 35.29 & 5.72 & 18.34 & 37.21 & 5.12 & 15.53 & 27.58 & 14.02 & 21.29 & 29.68 & 12.46 & 26.86 & 45.91 & 61.65 \\
            \hspace{1.6em}RECOMP & N/A & 8.14 & 19.06 & 41.02 & 7.43 & 22.13 & 41.45 & N/A & N/A & N/A & N/A & N/A & N/A & 18.51 & 35.76 & 56.40 & N/A \\
            \hspace{1.6em}ICAE & 4$\times$ & 14.43 & 25.99 & 37.83 & 9.53 & 25.03 & 38.12 & 9.20 & 22.36 & 32.88 & 26.84 & 33.80 & 36.71 & 24.77 & 40.06 & 53.82 & 62.70 \\
            \hspace{1.6em}xRAG & 164$\times$ & 3.02 & 12.45 & 40.83 & 5.37 & 17.92 & 40.28 & 2.95 & 14.22 & 34.88 & 10.06 & 18.30 & 36.66 & 8.45 & 22.52 & 45.65 & 63.00 \\
            \hspace{1.6em}COCOM & 128$\times$ & 31.86 & 40.15 & 42.44 & 13.75 & 30.68 & 37.58 & 20.08 & 32.19 & 35.94 & 21.72 & 26.21 & 25.32 & 25.90 & 35.36 & 34.00 & 60.87 \\
            \hspace{1.6em}PISCO & 64$\times$ & 0.06 & 9.49 & 39.87 & 0.80 & 9.11 & 42.11 & 0.20 & 10.24 & 31.57 & 0.44 & 13.28 & \cellcolor{gray!15}38.38 & 0.95 & 13.08 & 54.18 & \cellcolor{orange!20}\textbf{65.78} \\
            \hspace{1.6em}\textbf{\system{}} & 82$\times$ & \cellcolor{orange!20}\textbf{41.80} & \cellcolor{gray!15}49.72 & \cellcolor{orange!20}\textbf{50.91} & \cellcolor{gray!15}35.26 & \cellcolor{gray!15}47.27 & \cellcolor{gray!15}45.96 & \cellcolor{orange!20}\textbf32.04 & \cellcolor{orange!20}\textbf{40.95} & \cellcolor{orange!20}\textbf{40.05} & \cellcolor{orange!20}\textbf{31.42} & \cellcolor{orange!20}\textbf{37.61} & 37.65 & \cellcolor{orange!20}\textbf{44.46} & \cellcolor{orange!20}\textbf{57.88} & \cellcolor{gray!15}57.49 & \cellcolor{gray!15}65.23 \\
            \multicolumn{18}{l}{\hspace{0.8em}\textit{\textbf{Top k=5}}} \\
            \hspace{1.6em}LLM (with RAG*) & -- & \underline{45.72} & \underline{54.49} & \underline{55.73} & 37.71 & \underline{50.19} & \underline{49.06} & \underline{35.68} & \underline{44.29} & \underline{42.28} & \underline{34.20} & 40.17 & \underline{41.05} & 57.22 & 74.36 & 69.96 & 65.10 \\
            \hspace{1.6em}\textbf{\system{}} & 82$\times$ & 44.27 & 52.85 & 54.20 & \underline{37.74} & 49.83 & 48.37 & 33.68 & 42.04 & 41.52 & 33.32 & \underline{40.51} & 39.56 & \underline{61.13} & \underline{76.16} & \underline{76.57} & \underline{66.50} \\
            \midrule[0.8pt]
    
            \multicolumn{18}{l}{\textbf{Qwen2.5-7B-Instruct}} \\
            \multicolumn{18}{l}{\hspace{0.8em}\textit{\textbf{Top k=1}}} \\
            \hspace{1.6em}LLM (w/o RAG) & -- & 7.78 & 17.18 & 31.51 & 6.96 & 19.46 & 28.60 & 6.69 & 20.26 & 32.74 & 10.57 & 15.84 & 18.16 & 15.73 & 23.79 & 24.63 & 57.93 \\
            \hspace{1.6em}LLM (with RAG) & -- & 15.01 & 24.96 & 38.92 & 9.63 & 24.13 & 37.68 & 9.06 & 19.29 & 30.80 & 25.22 & 33.19 & \cellcolor{orange!20}\textbf{38.09} & 36.96 & 49.95 & \cellcolor{orange!20}\textbf{57.17} & 62.30 \\
            \hspace{1.6em}LLM (with RAG*) & -- & \cellcolor{orange!20}\textbf{39.31} & \cellcolor{orange!20}\textbf{48.15} & \cellcolor{orange!20}\textbf{49.91} & \cellcolor{gray!15}26.49 & \cellcolor{gray!15}41.12 & \cellcolor{gray!15}41.07 & \cellcolor{orange!20}\textbf{28.10} & \cellcolor{orange!20}\textbf{37.92} & \cellcolor{orange!20}\textbf{38.16} & \cellcolor{gray!15}29.33 & \cellcolor{gray!15}35.77 & 35.13 & \cellcolor{orange!20}\textbf{43.19} & \cellcolor{gray!15}52.61 & 51.71 & \cellcolor{gray!15}62.60 \\
            \hspace{1.6em}\textbf{\system{}} & 82$\times$ & \cellcolor{gray!15}38.58 & \cellcolor{gray!15}47.02 & \cellcolor{gray!15}49.02 & \cellcolor{orange!20}\textbf{30.47} & \cellcolor{orange!20}\textbf{42.43} & \cellcolor{orange!20}\textbf{42.42} & \cellcolor{gray!15}27.24 & \cellcolor{gray!15}36.94 & \cellcolor{gray!15}38.00 & \cellcolor{orange!20}\textbf{31.20} & \cellcolor{orange!20}\textbf{37.71} & \cellcolor{gray!15}37.04 & \cellcolor{gray!15}42.35 & \cellcolor{orange!20}\textbf{55.34} & \cellcolor{gray!15}55.91 & \cellcolor{orange!20}\textbf{67.44} \\
            \multicolumn{18}{l}{\hspace{0.8em}\textit{\textbf{Top k=5}}} \\
            \hspace{1.6em}LLM (with RAG*) & -- & \underline{42.76} & \underline{51.94} & \underline{54.03} & 28.95 & 44.76 & \underline{46.32} & 28.69 & 38.43 & 39.38 & \underline{32.28} & 38.52 & 37.92 & 56.26 & 72.77 & 74.94 & 64.60 \\
            \hspace{1.6em}\textbf{\system{}} & 82$\times$ & 41.86 & 50.94 & 52.22 & \underline{33.34} & \underline{45.56} & 44.69 & \underline{29.72} & \underline{39.18} & \underline{39.93} & 32.23 & \underline{38.71} & \underline{39.02} & \underline{59.14} & \underline{74.38} & \underline{75.43} & \underline{65.24} \\
            \bottomrule
            
        \end{tabular}
    }
    \par\smallskip
    \begin{flushleft}
        \footnotesize \textbf{Note:} RECOMP yields dynamic compression rate due to sentence-level selection. It's dataset-specific checkpoints on Web Questions, PopQA and FactKG are also not available.
    \end{flushleft}
\end{table*}

\subsubsection{Stage2: Training Information-utilizing Generation.}
Once the selector and projector have been trained, we focus on training the generator LLM to effectively utilize the inserted compressed embeddings for final generation.

Specifically, we use data from the training set of multiple public QA datasets $(q_i, a_i)$, obtain their retrieved documents $d_i$, feed the pair $(q_i, d_i)$ through the selector, projector, and the generator. Similar to Stage 1 training, we calculate the Next Token Prediction loss between the generated answer and the true answer $a_i$, and use this loss to update the generator. By minimizing this loss, we enable the generator to better utilize the embeddings provided by the selector and projector, thereby generating accurate answers. Prompts used in both training stages can be found at Appendix~\ref{appendix_j}.

\section{Experiment}
\subsection{Experimental settings}
\subsubsection{Datasets}
We conduct experiments on: (1) Open-Domain Question Answering (ODQA), including Natural Questions \cite{NQ}, TriviaQA \cite{TriviaQA}, Web Questions \cite{WebQA} and PopQA \cite{PopQA}. (2) Multihop Question Answering, represented by HotpotQA \cite{HotpotQA}. (3) Fact checking, represented by FactKG \cite{FactKG}.

\subsubsection{Metrics}
Following previous settings \cite{xRAG,COCOM}, we adopt Exact Match (EM) and F1-score (F1). In addition, we employ LLM-as-a-judge (LLM) using Qwen3-30B-A3B-Instruct. This offers greater robustness when answers are \emph{semantically} correct. Detailed prompt in Appendix~\ref{appendix_j}. For the FactKG dataset, we use Accuracy(Acc) only\cite{FactAcc}. Higher EM, F1, LLM and Acc indicate better performance.

\subsubsection{Implementation Details}
\system{} applies \emph{Qwen3-Embedding-0.6B} as the selector backbone, one-layer MLP as the projector, and \emph{Mistral-7B-Instruct-0.2}/\emph{Qwen2.5-7B-Instruct} as the generator backbone. We set $p=8$ and $n=2$. We define the compression rate as the ratio of the compressed content length to the number of compressed tokens. More details are provided in Appendix~\ref{appendix_b}.

\subsection{Baselines}
We evaluate \system{} against non-compression, representative hard compression baselines and soft compression baselines. 

Non-compression Baselines include:

\noindent$\bullet$~\textbf{LLM (w/o RAG)}: relies on internal knowledge to respond.

\noindent$\bullet$~\textbf{LLM (with RAG)}: standard RAG pipeline using the off-the-shelf frozen LLM for direct inference.

\noindent$\bullet$~\textbf{LLM (with RAG*)}: standard RAG pipeline fine-tuned on the same dataset as \system{}'s stage2 to ensure a fair comparison, serving as the strongest uncompressed baseline.

Hard compression-based methods include:

\noindent$\bullet$~\textbf{LLMLingua-2\cite{LLMLingua-2}}: Token-level classifier for hard compression.

\noindent$\bullet$~\textbf{RECOMP}\cite{RECOMP}: Sentence-level re-ranking for hard compression.

Soft compression-based methods include:

\noindent$\bullet$~\textbf{ICAE}\cite{ICAE}: In-context document auto-encoder with a fixed compression ratio.

\noindent$\bullet$~\textbf{xRAG}\cite{xRAG}: compresses a single document into a single token embedding for subsequent generation.

\noindent$\bullet$~\textbf{COCOM}\cite{COCOM}: an unified model employs auto-encoding to compress context into embeddings.

\noindent$\bullet$~\textbf{PISCO}\cite{PISCO}: applies sequence-level knowledge distillation for context compression without annotated data.

\subsection{Experimental Results}
\subsubsection{Performance}
Table~\ref{tab:performance} presents the main results. \system{} consistently outperforms all compression methods and remains competitive with non-compression baselines. Extending retrieval to top-5 further improves performance on multi-hop QA datasets like HotpotQA, showing that multiple compressed embeddings capture complementary evidence without introducing redundancy.

A similar trend holds for Qwen2.5-7B backbone. These results confirm that query-conditioned selection preserves critical reasoning cues even at extreme compression. Additionally, we provide the case study in Appendix~\ref{appendix_i}. 

\subsubsection{Efficiency}\label{5.3.2}
We evaluate efficiency with: Total Inference Latency (TIL), measuring end-to-end latency; GFLOPs, approximating the computational cost; and Time to First Token (TTFT), latency to generate the first output token.

As shown in Table~\ref{tab:efficiency}, \system{} achieves the best efficiency. Compared to non-compression baselines, both GFLOPs and TTFT is significantly lowered due to reduced generator context length. Compared to compression baselines, \system{} also demonstrates superior efficiency due to a lightweight selector. Under top-5, \system{} shows only slightly decreased efficiency, while the baseline’s runtime increases sharply, indicating that selective compression scales efficiently with different top-k settings. Efficiency evaluation across all datasets are presented in the Appendix~\ref{appendix_d}.
\begin{table}[h]
    \renewcommand{\arraystretch}{0.8}
    \centering
    \caption{Efficiency comparison on the Mistral with respect to LLM (with RAG*). \textcolor{green!60!black}{Lower values} indicate better efficiency.}
    \label{tab:efficiency}
    \resizebox{\columnwidth}{!}{
        \begin{tabular}{l cc cc cc}
            \toprule
            \multirow{2}{*}{\textbf{Method}} & 
            \multicolumn{2}{c}{\textbf{TIL (ms)}} & 
            \multicolumn{2}{c}{\textbf{GFLOPs}} &
            \multicolumn{2}{c}{\textbf{TTFT (ms)}} \\
            \cmidrule(lr){2-3} \cmidrule(lr){4-5} \cmidrule(lr){6-7}
            & \textbf{NQ} & \textbf{HotpotQA} & 
              \textbf{NQ} & \textbf{HotpotQA} &
              \textbf{NQ} & \textbf{HotpotQA} \\
            \midrule
            \multicolumn{7}{l}{\textit{\textbf{Top k=1}}} \\
            \hspace{0.8em}LLM (w/o RAG)        & 1096   & 1073   & 1188    & 1270    & 47   & 47   \\
            \hspace{0.8em}LLM (with RAG)       & 1092   & 1006   & 3186    & 2694    & 60   & 57   \\
            \hspace{0.8em}LLM (with RAG*)      & 1396   & 1024   & 3001    & 2570    & 69   & 63   \\
            \hspace{0.8em}LLMLingua-2              
                & 921~\textcolor{green!60!black}{(-34.0\%)} 
                & 716~\textcolor{green!60!black}{(-30.1\%)} 
                & 2431~\textcolor{green!60!black}{(-19.0\%)} 
                & 2082~\textcolor{green!60!black}{(-19.0\%)} 
                & 84~\textcolor{red!70!black}{(+21.7\%)} 
                & 76~\textcolor{red!70!black}{(+20.6\%)} \\
            \hspace{0.8em}RECOMP               
                & 754~\textcolor{green!60!black}{(-46.0\%)} 
                & 563~\textcolor{green!60!black}{(-45.0\%)} 
                & 2101~\textcolor{green!60!black}{(-30.0\%)} 
                & 1799~\textcolor{green!60!black}{(-30.0\%)} 
                & 98~\textcolor{red!70!black}{(+42.0\%)} 
                & 89~\textcolor{red!70!black}{(+41.3\%)} \\
            \hspace{0.8em}ICAE               
                & 841~\textcolor{green!60!black}{(-39.8\%)} 
                & 738~\textcolor{green!60!black}{(-28.0\%)} 
                & 3731~\textcolor{red!70!black}{(+24.3\%)} 
                & 3293~\textcolor{red!70!black}{(+28.1\%)} 
                & 169~\textcolor{red!70!black}{(+145.0\%)} 
                & 164~\textcolor{red!70!black}{(+160.3\%)} \\
            \hspace{0.8em}xRAG               
                & 1295~\textcolor{green!60!black}{(-7.2\%)} 
                & 1188~\textcolor{red!70!black}{(+16.0\%)} 
                & 1317~\textcolor{green!60!black}{(-56.1\%)} 
                & 1158~\textcolor{green!40!black}{(\textbf{-54.9\%})} 
                & 70~\textcolor{red!70!black}{(+1.4\%)} 
                & 71~\textcolor{red!70!black}{(+12.7\%)} \\
            \hspace{0.8em}COCOM              
                & 488~\textcolor{green!40!black}{(\textbf{-65.0\%})} 
                & 541~\textcolor{green!60!black}{(-47.2\%)} 
                & 2346~\textcolor{green!60!black}{(-21.8\%)} 
                & 1905~\textcolor{green!60!black}{(-25.9\%)} 
                & 138~\textcolor{red!70!black}{(+100.0\%)} 
                & 138~\textcolor{red!70!black}{(+119.0\%)} \\
            \hspace{0.8em}PISCO              
                & 502~\textcolor{green!60!black}{(-64.0\%)}
                & 553~\textcolor{green!60!black}{(-46.0\%)} 
                & 2251~\textcolor{green!60!black}{(-25.0\%)} 
                & 1927~\textcolor{green!60!black}{(-25.0\%)} 
                & 142~\textcolor{red!70!black}{(+105.8\%)} 
                & 141~\textcolor{red!70!black}{(+123.8\%)} \\
            \hspace{0.8em}\textbf{\system{}} 
                & 535~\textcolor{green!60!black}{(-61.7\%)} 
                & 496~\textcolor{green!40!black}{(\textbf{-51.6\%})} 
                & 1166~\textcolor{green!40!black}{(\textbf{-61.1\%})} 
                & 1271~\textcolor{green!60!black}{(-50.6\%)} 
                & 49~\textcolor{green!40!black}{(\textbf{-29.0\%})} 
                & 49~\textcolor{green!40!black}{(\textbf{-22.2\%})} \\

            \multicolumn{7}{l}{\textit{\textbf{Top k=5}}} \\
            \hspace{0.8em}LLM (with RAG*)      & 1537   & 1459   & 10811   & 4038    & 165  & 80   \\
            \hspace{0.8em}\textbf{\system{}}   
                & 590~\textcolor{green!60!black}{(\textbf{-61.6\%})} 
                & 505~\textcolor{green!60!black}{(\textbf{-65.4\%})} 
                & 1664~\textcolor{green!60!black}{(\textbf{-84.6\%})} 
                & 1416~\textcolor{green!60!black}{(\textbf{-64.9\%})} 
                & 72~\textcolor{green!60!black}{(\textbf{-56.4\%})} 
                & 53~\textcolor{green!60!black}{(\textbf{-33.8\%)}} \\
            \bottomrule
        \end{tabular}%
    }
\end{table}

\section{Extended Experiments}
In this section, we evaluate how \system{} addresses the limitations of full-compression methods (\S\ref{6.1}), \system{}'s generalization and scalability (\S\ref{6.2}), and ablation on training strategies (\S\ref{6.3}). We also present ablations on projector and normalization (Appendix~\ref{appendix_e}), sensitivity to $p$ and $n$ (Appendix~\ref{appendix_f}) and robustness under poor retrieval (Appendix~\ref{appendix_g}).

\subsection{Extended Analysis}\label{6.1}
\subsubsection{\system{} in instruction following}
\begin{figure}[h]
  \centering
  \begin{subfigure}{0.5\columnwidth}
    \includegraphics[width=\linewidth]{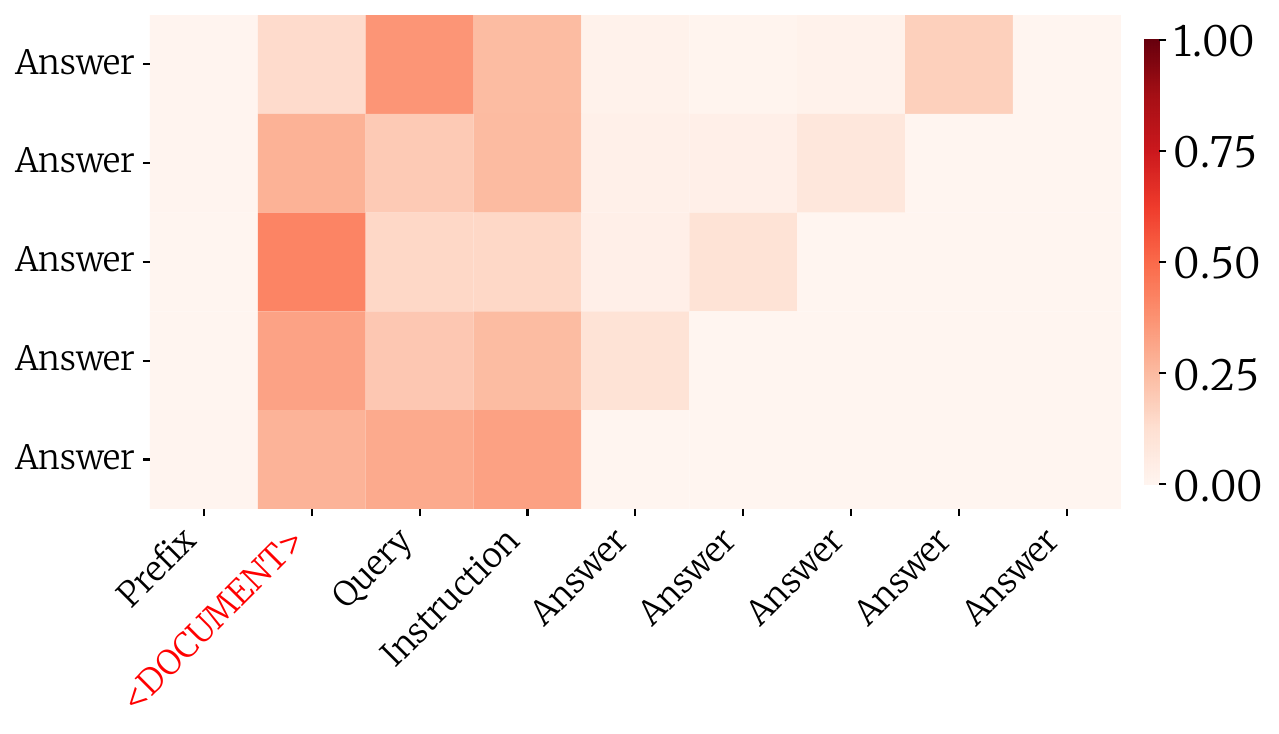}\label{fig:extend_our_re}
  \end{subfigure}
  \begin{subfigure}{0.449\columnwidth}\includegraphics[width=\linewidth]{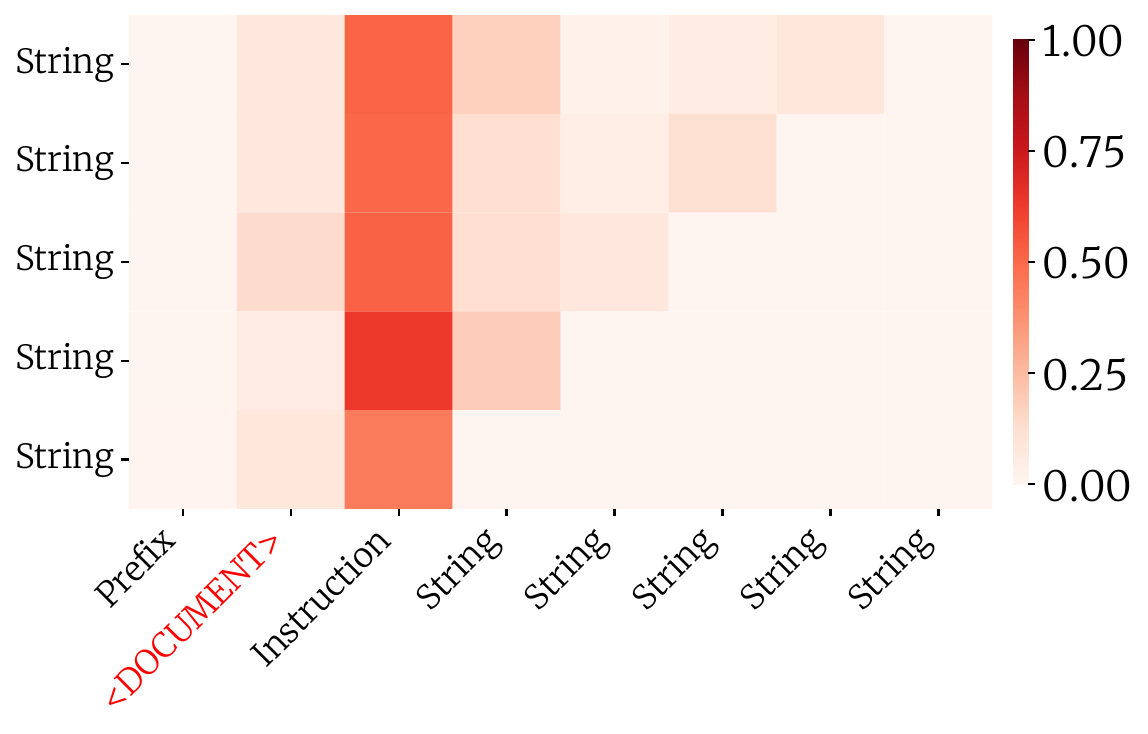}\label{fig:extend_our_mis}
  \end{subfigure}
  \caption{Attention heatmap of \system{}'s generator under QA (left) and instruction-following (right) prompts.}
  \label{fig:extend_our_fea}
\end{figure}

Under the same setup as \S\ref{observations_3.2}, we visualize attention of the generator in \system{} during decoding. As shown in Figure~\ref{fig:extend_our_fea}, when doint reconstruction, attention peaks correctly occur at the instruction, query, and <DOCUMENT> area. When doing instruction-following, attention shifts almost entirely to the instruction, assigning negligible mass to <DOCUMENT> while still producing correct responses. This behavior contrasts with full-compression methods, demonstrating the intact instruction-following capability of \system{}.
\subsubsection{\system{} in information selection}
\begin{figure}[h]
  \centering
  \includegraphics[width=0.9\columnwidth]{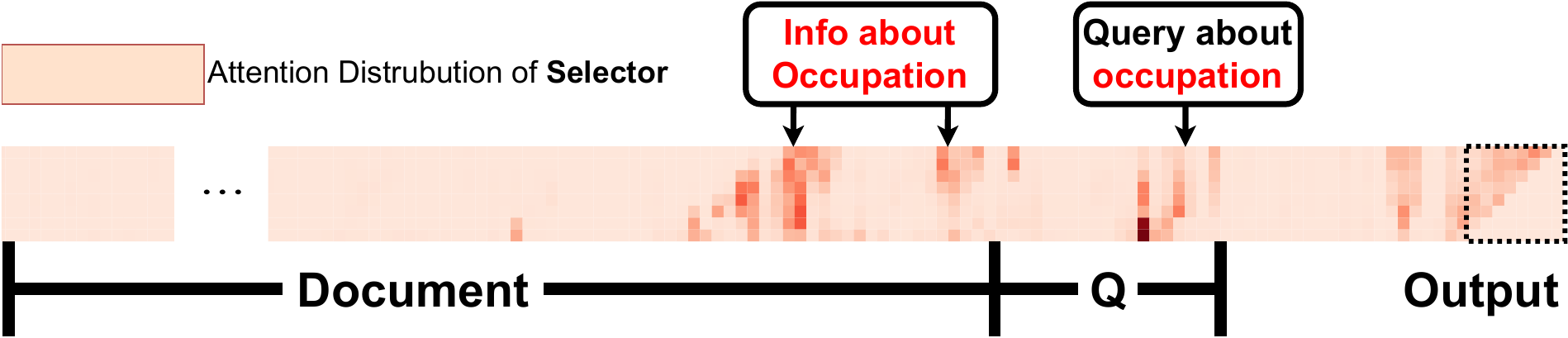}
  \caption{Attention of \system{}'s selector under QA prompt.}
  \label{fig:extend_our_nece}
\end{figure}

Under the same setup as \S\ref{observations_3.3}, we visualize the selector’s attention while generating the \texttt{<ENCODE>} embeddings. As shown in Figure~\ref{fig:extend_our_nece}, the heatmap shows sharp, localized peaks on query and query-relevant document regions, while irrelevant regions are ignored. This pattern confirms that our selector acts as a query-conditioned information selector that yields cleaner, redundancy-free embeddings for the generator.

\subsection{Generalization and Scalability}\label{6.2}
We evaluate \system{} across varying model scales and architectures (Table~\ref{tab:scalability}).
Tested on Qwen2.5-3B and Llama-2-13B generators, \system{} consistently demonstrates performance parallel to RAG baselines. This confirms that query-conditioned selection effectively preserves reasoning cues regardless of generator model family and capacity.
Using different backbones of selectors also maintains competitive performance, despite expected performance drop due to reduction in the selector size. These results confirm that the \system{} framework can effectivelly generalize and scale to different backbone sizes and architectures.
\begin{table}[h]
    \centering
    \renewcommand{\arraystretch}{0.8}
    \small 
    \setlength{\tabcolsep}{3.5pt} 
    \caption{Robustness over generator and selector backbones, including sizes and architectures.} 
    \label{tab:scalability}
    \resizebox{\columnwidth}{!}{
    \begin{tabular}{ll ccc ccc}
        \toprule
        \multirow{2}{*}{\textbf{Method}} & \multirow{2}{*}{\textbf{Backbone}} 
        & \multicolumn{3}{c}{\textbf{NQ}} 
        & \multicolumn{3}{c}{\textbf{HotpotQA}} \\
        \cmidrule(lr){3-5} \cmidrule(lr){6-8}
        & & \textbf{EM} & \textbf{F1} & \textbf{LLM} 
        & \textbf{EM} & \textbf{F1} & \textbf{LLM} \\
        
        \midrule

        \multicolumn{8}{l}{\textit{\textbf{Generator} (Fixed Selector: Qwen3-0.6B)}} \\
        \midrule
        LLM w/ RAG* & \multirow{2}{*}{Qwen2.5-3B} & 33.96 & 42.77 & 44.11 & 35.54 & 48.40 & 52.30 \\
        \system{}     &                             & 32.86 & 41.01 & 42.19 & 40.34 & 53.31 & 52.81 \\
        \cmidrule{1-8}
        LLM w/ RAG* & \multirow{2}{*}{Llama2-13B} & 46.00 & 54.63 & 56.08 & 42.85 & 57.11 & 57.16 \\
        \system{}     &                             & 44.99 & 53.59 & 54.99 & 46.21 & 59.59 & 59.78 \\
        
        \midrule
        
        \multicolumn{8}{l}{\textit{\textbf{Selector} (Fixed Generator: Mistral-7B)}} \\
        \midrule
        \system{} & GPT2-small  & 39.83 & 48.61 & 50.88 & 42.43 & 56.09 & 55.33 \\
        \system{} & SmolLM-360m & 42.85 & 51.65 & 53.38 & 47.54 & 60.95 & 61.03 \\
        \system{} & Qwen3-Embedding-0.6B   & 41.80 & 49.72 & 50.91 & 44.46 & 57.88 & 57.49 \\ 
        \bottomrule
    \end{tabular}
    }
\end{table}
\subsection{Ablation on Training Strategy}\label{6.3}
We ablate four training strategy: (I) w/o stage1 ft,  which removes the first-stage selector training; (II) w/o stage2 ft, which skips the second-stage generator fine-tuning; (III) stage2 (selector tuned), which jointly tunes the selector and generator in stage two; and (IV) Ours, which integrates both stages.

\begin{figure}[h]
  \centering
  \includegraphics[width=0.6\columnwidth]{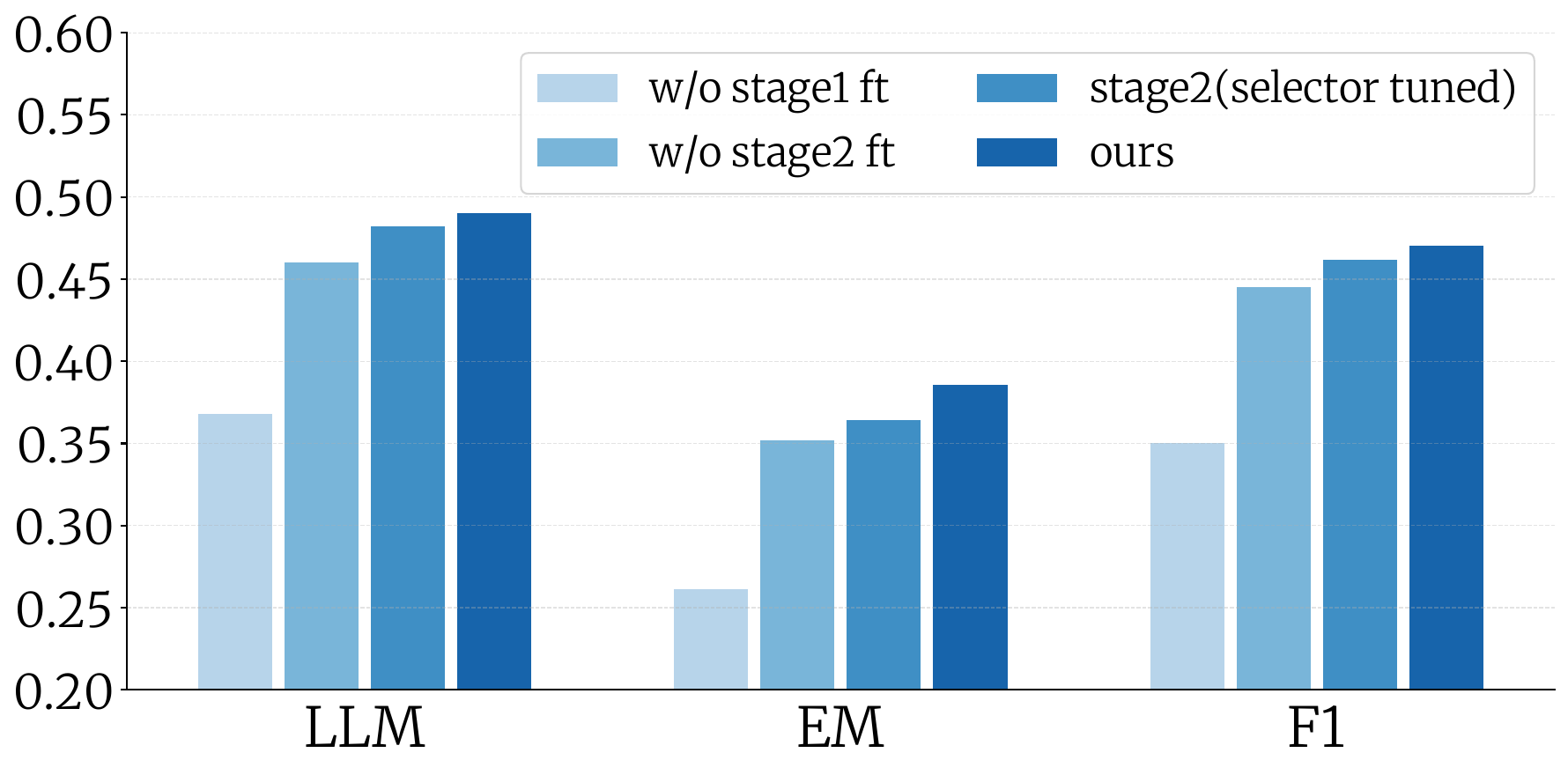}
  \caption{Ablation study on different training strategy.}
  \label{fig:ablation_train}
\end{figure}

As shown in Figure~\ref{fig:ablation_train}, both stages contribute positively. Without stage1 training (w/o stage1 ft), performance on three metrics drops markedly, indicating that the selector requires explicit supervision to extract query-conditioned information. Removing stage2 fine-tuning (w/o stage2 ft) leads to weaker generator adaptation, reflecting the importance of teaching the generator how to utilize compressed embeddings. Stage2 (selector tuned) improves over either stage alone, but still lags behind our two-stage process.

We also validates the effectiveness of curriculum learning. Without curriculum learning, F1-Score drops by 0.2\% to 0.5\% across all datasets. Curriculum learning stabilizes optimization, allowing the model to progressively develop robust extraction capabilities.
\section{Conclusion}
This paper revisited the full-compression paradigm of soft compression RAG methods and demonstrates its infeasibility and non-necessity. We introduced \system{} that shifts the role of encoder as a query-conditioned selector to preserve critical evidence under high compression. We carried out a comprehensive data synthesis framework to curate a massive and diverse dataset to train the selector. We validated that shifting to an autoregressive selection paradigm effectively prevents information dilution and aligns compressed representations with downstream generation behaviors. Experiments confirm that \system{} achieves state-of-the-art performance while significantly reducing computation and latency. We open-source our code\footnote{\url{https://github.com/yhliu7458/SeleCom}}, data\footnote{\url{https://huggingface.co/datasets/Ryan7458/SeleCom_Training}} and model checkpoints\footnote{\url{https://huggingface.co/Ryan7458/Selecom}}.

%%\begin{acks}

%%\end{acks}

\bibliographystyle{ACM-Reference-Format}
\balance
\bibliography{sample-base}

\appendix
\section{Analysis Details}\label{appendix_a}
\subsection{Infeasibility of Full Compression}\label{appendix_a1}
\subsubsection{Detailed theory}
\textbf{(1)} The full-compression task first pushes the compressed embeddings $Z$ towards a "positive half-space" of the LLM's token embeddings, causing direction similarity that results in positive attention scores.

Under a hyperspherical view, each query vector in decoder's attention layer can be decomposed into a unit direction (on-sphere) and a radius (norm). Modern decoder-only LLM representations are markedly anisotropic—mass concentrates in a narrow cone and a few high-variance "outlier" features dominate the similarity structure~\cite{ana1_1,ana1_2,ana1_3,ana1_4}. This implies the existence of a "positive half-space" direction \(v\) such that the vast majority of tokens \(E=\{e_w\}_{w\in\mathcal{V}}\subset\mathbb{R}^d\) lie in its positive half-space: $\Pr_{w\sim\mathcal{V}}\big[\cos(e_w,\,v)> 0\big]\approx 1.$

During the full-compression training task, the model's desired output fully rely on the $Z$, which forces the attention score on $Z$ to consistently increase. This forces $Z$ to gradually align toward this direction \(v\), "hacking" the generator attention layer to produce positive attention scores on these embeddings.

This sets up the next cause: \textbf{(2)} Once the embedding direction is aligned, the everlasting training pressure generated by the full-compression task will push the norm \(R_Z\) of the $Z$ higher and higher, systematically amplifying attention logits that causes polarized attention toward $Z$. 

Specifically, the compressed embeddings $Z$ must contain as much information as possible from the retrieved document for the downstream LLM to correctly reconstruct the document. From the information theory perspective, we refer to the Shannon–Hartley theorem~\cite{ana2}: $C=Blog_2(1+\frac{S}{N})$. where $C$ is the expressing capacity of the $Z$, $B$ is the channel bandwidth determined by the LLM, and $N$ is the noise power. With the decoder frozen and noise fixed, $Z$'s expressing capacity monotonically increases with $S$, the signal power, which corresponds to the the embedding norm \(R_Z\). Thus, training effectively inflates the embedding’s magnitude to transmit more information reliably across layers and tokens.

Furthermore, the strong pressure to raise attention scores also inherently pushes up \(R_Z\), which serves as the \emph{key vector} of all decoded tokens in the attention mechanism.

\textbf{(3)} While the full-compression task does not enforce instruction-following abilities, attention scores to other tokens remain low. Consequently, once the direction of the $Z$ is correct, and $R_Z$ grows high, softmax normalization rapidly polarizes attention toward $Z$. This drives attention to collapse onto the compressed document embeddings across layers. When prompts include special instructions or auxiliary information, the LLM simply ignores them and keeps their pre-defined behaviors as if it sees only the compressed embeddings. Consequently, the model’s ability to analyze the query and extract supporting evidence deteriorates, which further reduces its performance on downstream tasks.

\subsubsection{Detailed experimental setup}
A pipeline of \emph{Qwen3-Embedding-0.6B} as the full-compression encoder, a frozen \emph{Qwen2.5-7B-Instruct} as the generator and a one-layer MLP as the projector is trained on the document reconstruction task with 10,000,000 random samples from the wikipedia dump. We then apply different prompts to test its query-conditioned generation ability.

\subsection{Non-necessity of Full Compression}\label{appendix_a2}
\subsubsection{Detailed theory}
Let \(Q\) be the query, \(D\) as the retrieved document, \(A\) as the answer, and \(E\!\subseteq\! D\) the necessary information, with \(N = D \setminus E\) as the set of noisy information. According to the definition of \(E\), we have:
$$
I(A;D \mid Q) \approx I(A;E \mid Q) + I(A;N \mid Q)=I(A;E \mid Q).
$$

Let \(Z = f(D)\) be the full-compression encoder's document embeddings, and S\(Z_E = f(E)\) and S\(Z_N = f(N)\) respectively be the necessary and noisy information compressed. \(Z\) is a capacity-limited and query-agnostic function of \((E,N)\), part of its bitrate is inevitably spent preserving information about \(N\), yielding:
$$
I(A;Z \mid Q) < I(A;Z_E \mid Q) + I(A;Z_N \mid Q) =I(A;Z_E \mid Q).
$$
Thus, full compression with no focus retains strictly less helpful information than compressing only the query-conditioned necessary information.

On the other hand, if we directly build a selector that given $Q$, it extracts and compresses only the necessary information as $Z_s=f_s(D)=f_s(E)$. Its superior capability than the full-compression encoder can be strictly proven:
$$
I(A;Z_s\mid Q) =I(A;Z_E\mid Q) > I(A;Z \mid Q).
$$
In this case, the selector’s document embeddings carries information that \emph{aligns with the query} and is only \emph{slightly less than} the baseline, while full compression is strictly worse due to capacity spent on \(N\) and lossy entanglement of \(E\).

\subsubsection{Detailed experimental setup}
The baseline RAG results are obtained with \emph{Qwen2.5-7B-Instruct}. The full-compression results are obtained with the trained pipeline during the infeasibility analysis. Under the same document, the baseline RAG is query-conditioned and the full-compression encoder is not.

\section{Main Experiment settings}\label{appendix_b}
All experiments are completed on a Ubuntu 22.04.1 LTS server with a Intel(R) Xeon(R) Gold 6530 CPU and 8$\times$NVIDIA GeForce RTX 4090 GPU (48GB) installed. We conduct stage1 training by 1 epoch, with batch size = 10 per device and learning rate = 0.00005. We perform stage2 training by 3 epoch, using LoRA r = 64, batch size = 3 per device and learning rate = 0.0001. For training time, Stage1 takes $\sim$ 40h, Stage2 takes $\sim$ 12h.

\section{Notations}\label{appendix_c}
Definitions of the notations are provided in Table~\ref{tab:notation}
\begin{table}[h]
    \centering
    \caption{Notations used in section Methodology.}
    \label{tab:notation}
    \renewcommand{\arraystretch}{0.8}
    \resizebox{0.85\columnwidth}{!}{
        \begin{tabular}{c|c}
        \toprule
        \textbf{Symbol} & \textbf{Description} \\
        \midrule
        $q_i, d_i, a_i$ & The $i$-th query, document, and answer triple. \\
        $k$ & Number of retrieved documents. \\
        $P_{select}, P_{gen}$ & Prompts for the Selector and Generator. \\
        $T_i$ & Token sequence for the $i$-th input. \\
        $p$ & Number of special $\langle En \rangle$ tokens for selection. \\
        $n$ & Number of embeddings per document. \\
        $H^{i}_{1:p}$ & Hidden vectors corresponding to $\langle En \rangle$ tokens. \\
        $E_i$ & Compressed embeddings set for document $d_i$. \\
        $\langle \mathrm{emb} \rangle$ & Special placeholder token. \\
        $S_i$ & Embedded sequence of $P_{gen}$ before inserting $E_i$. \\
        $\hat S_i$ & Final embedded sequence after inserting $E_i$. \\
        $\mathrm{Tok}(\cdot)$ & Tokenization function. \\
        $\mathrm{Emb}(\cdot)$ & Token embedding function. \\
        $\mathrm{Insert}(\cdot)$ & Function inserting $E_i$ into placeholder positions. \\
        $\text{Select}(\cdot)$ & The Selector model. \\
        $\text{Proj}(\cdot)$ & The Projector model. \\
        $\text{Split}_n(\cdot)$ & Function splitting a vector into $n$ sub-vectors. \\
        $\mathcal{L}_{CE}$ & Cross-entropy loss function. \\
        \bottomrule
        \end{tabular}
    }
\end{table}

\section{Comprehensive Experimental results}\label{appendix_d}
We present the efficiency results on all datasets in Table~\ref{tab:efficiency_total}.

\section{Ablation on Projector and Normalization}\label{appendix_e}
We evaluate four projector designs: (1) Transformer-based projector; (2) Token-wise MLP followed by learnable aggregation; (3) Flatten two-layer MLP with LayerNorms; (4) our one-layer MLP.

For normalization, we evaluate HardNorm-$n$, which enforces strict normalization to the norm value of $n$; SoftNorm-$n$, which rescales embeddings by a smooth norm-target function; and our NoNorm, which imposes no explicit constraint.

We also evaluate choices for compressed embedding normalization: HardNorm-$n$ enforcing strict normalization to the norm value of $n$; SoftNorm-$n$ applying smooth rescaling target; our NoNorm, with no normalization at all.

\begin{figure}[h]
  \centering
  \begin{subfigure}{0.48\columnwidth}
    \includegraphics[width=\linewidth]{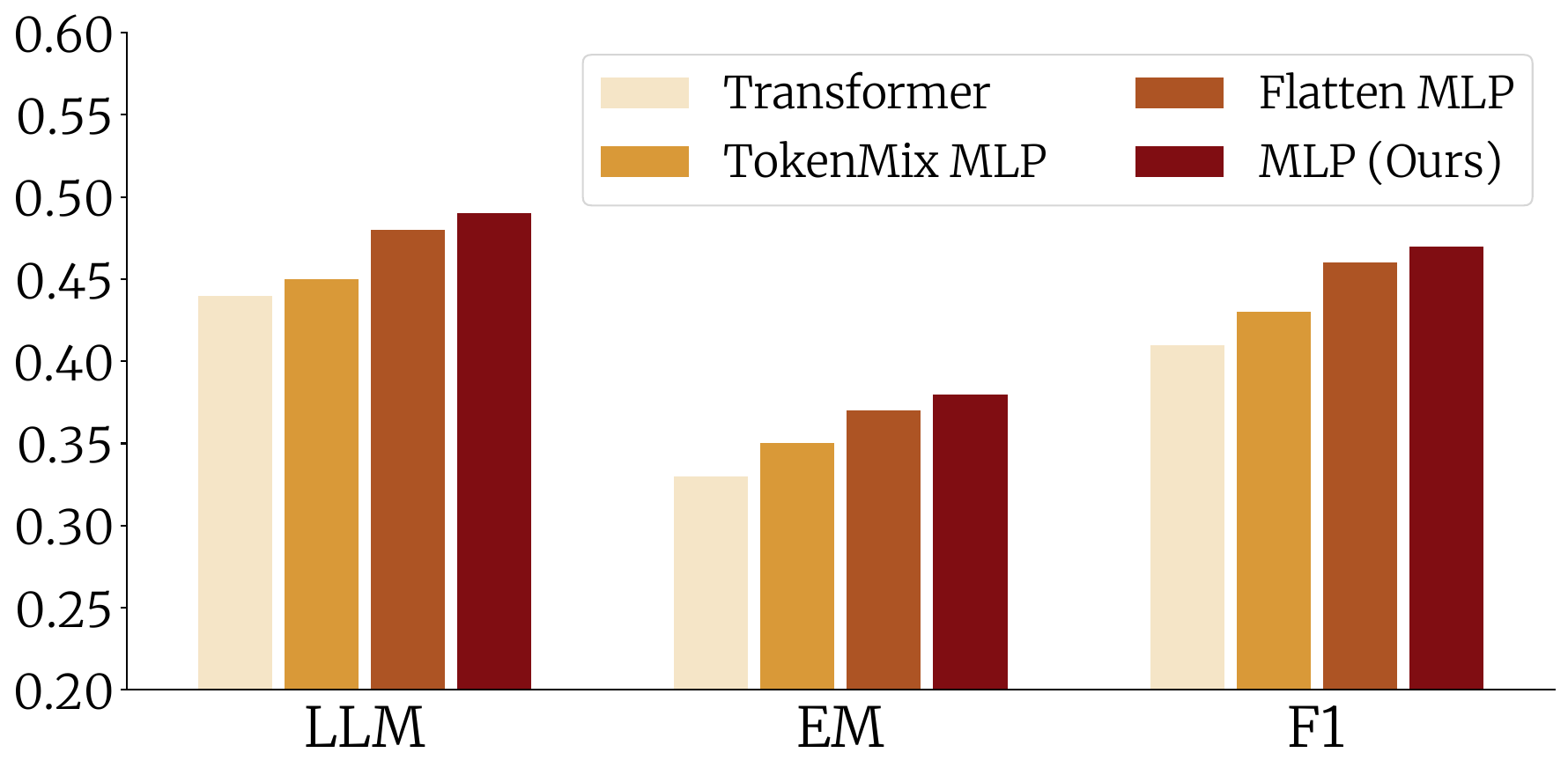}
    \caption{Proj. Performance}
    \label{fig:ablation_emb_proc_a}
  \end{subfigure}
  \hfill
  \begin{subfigure}{0.48\columnwidth}
    \includegraphics[width=\linewidth]{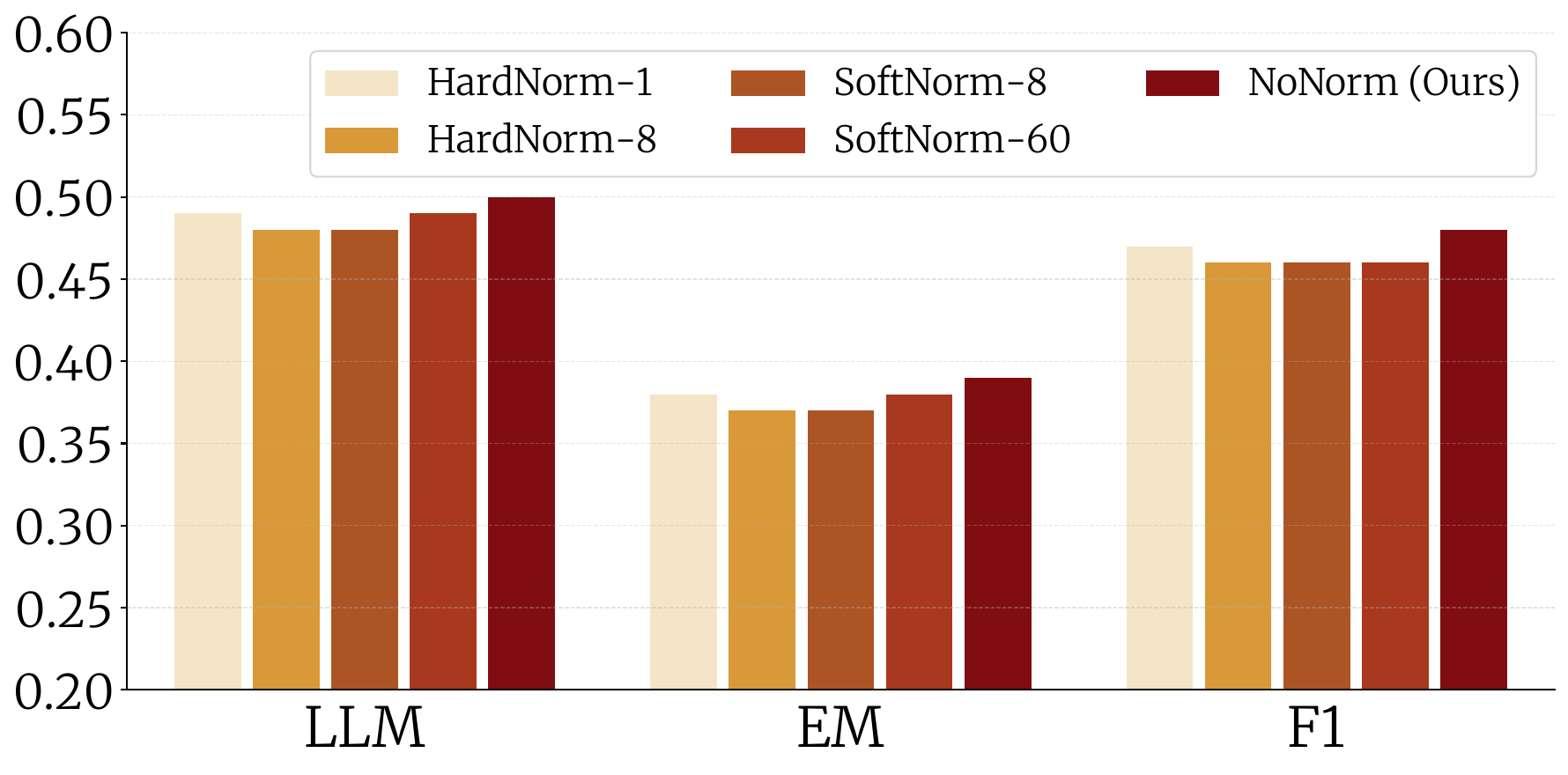}
    \caption{Norm. Performance}
    \label{fig:ablation_emb_proc_b}
  \end{subfigure}

  \begin{subfigure}{0.48\columnwidth}
    \includegraphics[width=\linewidth]{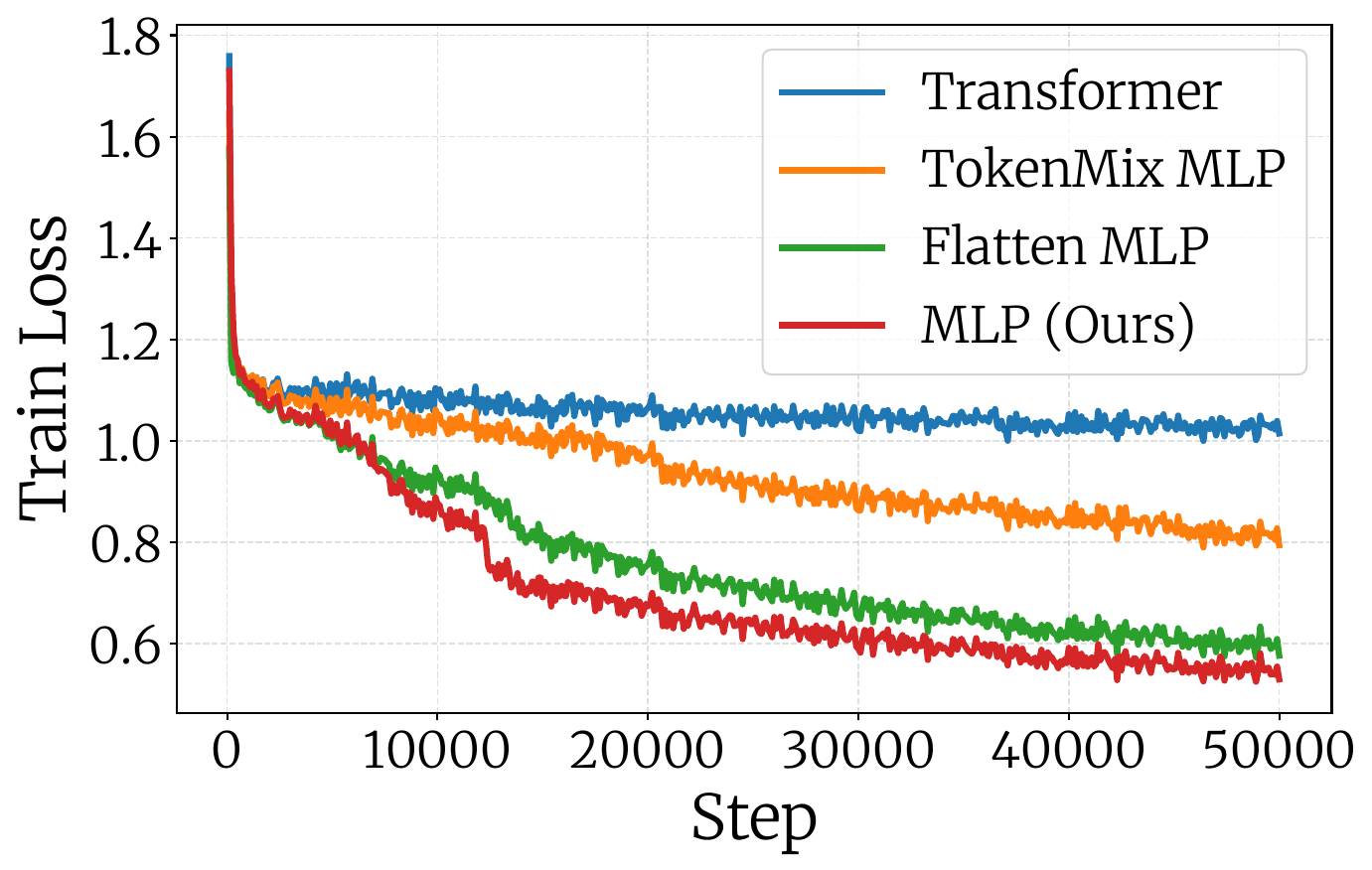}
    \caption{Proj. Train Loss}
    \label{fig:ablation_emb_proc_c}
  \end{subfigure}
  \hfill
  \begin{subfigure}{0.48\columnwidth}
    \includegraphics[width=\linewidth]{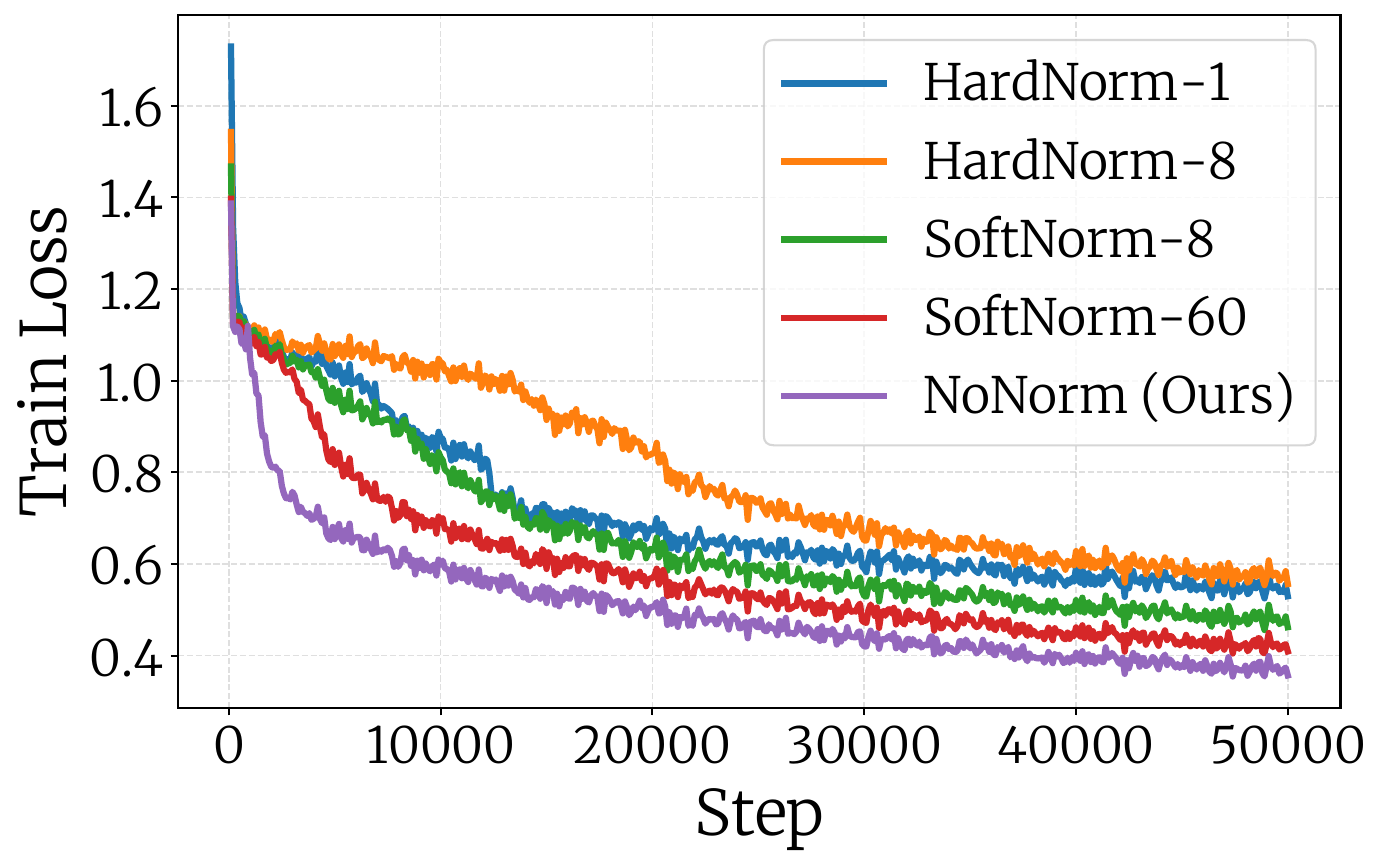}
    \caption{Norm. Train Loss}
    \label{fig:ablation_emb_proc_d}
  \end{subfigure}

  \caption{Ablation on normalization and projector design.}
  \label{fig:ablation_emb_proc}
\end{figure}

Results are shown in Figure~\ref{fig:ablation_emb_proc}. For projectors, our lightweight MLP achieves the best performance and fastest convergence. For normalization, both HardNorm and SoftNorm lead to suboptimal performance. These findings strengthen our design choices.

\section{Sensitivity to $p$ and $n$}\label{appendix_f}
We examine the impact of number of selector tokens ($p$) and compressed embeddings ($n$). As shown in Table~\ref{tab:pn}, \system{} remains robust across configurations.
\begin{table}[h]
    \centering
    \small 
    \renewcommand{\arraystretch}{0.6} 
    
    \setlength{\tabcolsep}{5pt}
    
    \caption{Sensitivity on $p$ and $n$.} 
    \label{tab:pn}
    \resizebox{0.75\columnwidth}{!}{
    \begin{tabular}{cc ccc ccc}
        \toprule
        \multicolumn{2}{c}{\textbf{Config.}} & 
        \multicolumn{3}{c}{\textbf{NQ}} & 
        \multicolumn{3}{c}{\textbf{HotpotQA}} \\
        
        \cmidrule(r){1-2} \cmidrule(lr){3-5} \cmidrule(l){6-8}
        
        $p$ & $n$ & 
        EM & F1 & LLM & 
        EM & F1 & LLM \\
        \midrule
        
        8 & 1 & 41.86 & 50.07 & 51.33 & 45.66 & 58.05 & 57.30 \\
        8 & 4 & 41.78 & 49.61 & 51.15 & 44.90 & 57.54 & 56.86 \\
        \midrule
        4 & 2 & 41.80 & 50.09 & 51.97 & 45.43 & 58.04 & 58.60 \\
        16 & 2 & 41.94 & 50.04 & 50.89 & 44.10 & 57.34 & 57.45 \\
        \midrule
        8 & 2 & 41.80 & 49.72 & 50.91 & 44.46 & 57.88 & 57.49 \\
        \bottomrule
    \end{tabular}
    }
\end{table}

\section{Robustness to Irrelevant Context}\label{appendix_g}
We evaluate robustness against retrieval failure by replacing top-$k$ documents with random corpus samples. As shown in Table~\ref{tab:irrelevant}, we report the F1 degradation ($\Delta$F1) compared to standard retrieval. \system{} demonstrates robustness parallel to the strongest non-compression baseline ($\max(\text{RAG}, \text{RAG}^*)$). Notably, on multi-hop QA (HotpotQA), SeleCom exhibits smaller performance drops, confirming that the query-conditioned selector effectively filters out noise that distracts the generator.
\begin{table}[h]
    \centering
    \small 
    \renewcommand{\arraystretch}{0.6} 
    \setlength{\tabcolsep}{3.5pt} 
    \caption{Performance drop ($\Delta$F1 $\downarrow$) under random documents.}
    \label{tab:irrelevant}
    \begin{tabular}{lcccccc}
        \toprule
        \textbf{Method} & \textbf{NQ} & \textbf{TriviaQA} & \textbf{WebQA} & \textbf{PopQA} & \textbf{HotpotQA} & \textbf{FactKG} \\
        \midrule
        Baseline & 3.6\% & 7.1\% & 1.1\% & 13.8\% & 15.7\% & 10.7\% \\
        \textbf{\system{}} & 3.7\% & 6.0\% & 1.0\% & 14.0\% & 11.1\% & 4.1\% \\
        \bottomrule
    \end{tabular}
\end{table}

\section{Human Verification on LLM Usage}\label{appendix_h}
For dataset synthesis, we randomly select 10 human experts to verify the quality of 1000 samples from the constructed dataset. They report quality of document, factuality and supportability of QA, and human alignment for difficulty labeling. For LLM-as-a-judge, we randomly select 6 human experts, each verifying 200 judgement samples for each evaluation dataset of \system{}. They report judgement consistency. Results are shown in Table~\ref{tab:human_eval}, which validates the quality of the LLM-synthetic dataset and the utility of LLM-as-a-judge.
\begin{table}[h]
    \centering
    \small
    \renewcommand{\arraystretch}{0.6} 
    \caption{Human verification results on LLM usage}
    \label{tab:human_eval}
        \begin{tabular}{llc}
            \toprule
            \textbf{Component} & \textbf{Criterion} & \textbf{Result (\%)} \\
            \midrule
            \multirow{4}{*}{Synthetic Data} 
             & Document Quality & 98.2 \\
             & Factuality & 99.5 \\
             & Supportability & 99.9 \\
             & Difficulty Label & 91.0 \\
            \midrule
            LLM-as-a-judge & Score Consistency & 97.5 \\
            \bottomrule
        \end{tabular}
\end{table}

\begin{table*}[h]
    \renewcommand{\arraystretch}{0.8}
    \centering
    \caption{Complete efficiency results on six datasets with three evaluation metrics.}
    \label{tab:efficiency_total}
    \resizebox{\textwidth}{!}{
    \begin{tabular}{lcccccccccccccccccc}
        \toprule
        \textbf{Models}
        & \multicolumn{3}{c}{\textbf{Natural Questions}}
        & \multicolumn{3}{c}{\textbf{TriviaQA}}
        & \multicolumn{3}{c}{\textbf{Web Questions}}
        & \multicolumn{3}{c}{\textbf{PopQA}}
        & \multicolumn{3}{c}{\textbf{HotpotQA}}
        & \multicolumn{3}{c}{\textbf{FactKG}} \\
        \cmidrule(lr){2-4}\cmidrule(lr){5-7}\cmidrule(lr){8-10}
        \cmidrule(lr){11-13}\cmidrule(lr){14-16}\cmidrule(lr){17-19}
        & TIL & GFLOPs & TTFT
        & TIL & GFLOPs & TTFT
        & TIL & GFLOPs & TTFT
        & TIL & GFLOPs & TTFT
        & TIL & GFLOPs & TTFT
        & TIL & GFLOPs & TTFT \\
        \midrule

        \multicolumn{19}{l}{\textbf{Mistral-7B-Instruct}} \\
        \multicolumn{19}{l}{\hspace{0.8em}\textit{\textbf{Top k=1}}} \\
        \hspace{1.6em}LLM (w/o RAG) & 1096 & 1188 & 47 & 941 & 1218 & 55 & 990 & 1144 & 47 & 953 & 1080 & 47 & 1072 & 1270 & 47 & 1087 & 1286 & 48 \\
        \hspace{1.6em}LLM (with RAG) & 1092 & 3186 & 60 & 956 & 3216 & 61 & 998 & 3043 & 59 & 810 & 2778 & 57 & 1006 & 2694 & 57 & 1162 & 3037 & 60 \\
        \hspace{1.6em}LLM (with RAG*) & 1396 & 3001 & 69 & 936 & 3101 & 73 & 1430 & 2879 & 69 & 881 & 2688 & 66 & 1024 & 2570 & 63 & 743 & 3002 & 68 \\
        \hspace{1.6em}LLMLingua-2
        & 921~\textcolor{green!60!black}{(-34.0\%)} 
        & 2431~\textcolor{green!60!black}{(-19.0\%)} 
        & 84~\textcolor{red!70!black}{(+21.7\%)} 
        & 655~\textcolor{green!60!black}{(-30.0\%)} 
        & 2512~\textcolor{green!60!black}{(-19.0\%)} 
        & 88~\textcolor{red!70!black}{(+20.5\%)} 
        & 986~\textcolor{green!60!black}{(-31.0\%)} 
        & 2332~\textcolor{green!60!black}{(-19.0\%)} 
        & 82~\textcolor{red!70!black}{(+18.8\%)} 
        & 617~\textcolor{green!60!black}{(-30.0\%)} 
        & 2177~\textcolor{green!60!black}{(-19.0\%)} 
        & 79~\textcolor{red!70!black}{(+19.7\%)} 
        & 716~\textcolor{green!60!black}{(-30.1\%)} 
        & 2082~\textcolor{green!60!black}{(-19.0\%)} 
        & 76~\textcolor{red!70!black}{(+20.6\%)} 
        & 520~\textcolor{green!60!black}{(-30.0\%)} 
        & 2432~\textcolor{green!60!black}{(-19.0\%)} 
        & 81~\textcolor{red!70!black}{(+19.1\%)} \\
        \hspace{1.6em}RECOMP
        & 754~\textcolor{green!60!black}{(-46.0\%)} 
        & 2101~\textcolor{green!60!black}{(-30.0\%)} 
        & 98~\textcolor{red!70!black}{(+42.0\%)} 
        & 515~\textcolor{green!60!black}{(-45.0\%)} 
        & 2171~\textcolor{green!60!black}{(-30.0\%)} 
        & 103~\textcolor{red!70!black}{(+41.1\%)} 
        & 786~\textcolor{green!60!black}{(-45.0\%)} 
        & 2015~\textcolor{green!60!black}{(-30.0\%)} 
        & 97~\textcolor{red!70!black}{(+40.6\%)} 
        & 485~\textcolor{green!60!black}{(-45.0\%)} 
        & 1882~\textcolor{green!60!black}{(-30.0\%)} 
        & 92~\textcolor{red!70!black}{(+39.4\%)} 
        & 563~\textcolor{green!60!black}{(-45.0\%)} 
        & 1799~\textcolor{green!60!black}{(-30.0\%)} 
        & 89~\textcolor{red!70!black}{(+41.3\%)} 
        & 409~\textcolor{green!60!black}{(-45.0\%)} 
        & 2101~\textcolor{green!60!black}{(-30.0\%)} 
        & 95~\textcolor{red!70!black}{(+39.7\%)} \\
        \hspace{1.6em}ICAE 
        & 841~\textcolor{green!60!black}{(-39.8\%)} 
        & 3731~\textcolor{red!70!black}{(+24.3\%)} 
        & 169~\textcolor{red!70!black}{(+145.0\%)} 
        & 615~\textcolor{green!60!black}{(-34.3\%)} 
        & 3812~\textcolor{red!70!black}{(+22.9\%)} 
        & 165~\textcolor{red!70!black}{(+126.0\%)} 
        & 929~\textcolor{green!60!black}{(-35.0\%)} 
        & 3683~\textcolor{red!70!black}{(+27.9\%)} 
        & 171~\textcolor{red!70!black}{(+147.8\%)} 
        & 699~\textcolor{green!60!black}{(-20.7\%)} 
        & 3186~\textcolor{red!70!black}{(+18.5\%)} 
        & 160~\textcolor{red!70!black}{(+142.4\%)} 
        & 738~\textcolor{green!60!black}{(-28.0\%)} 
        & 3293~\textcolor{red!70!black}{(+28.1\%)} 
        & 164~\textcolor{red!70!black}{(+160.3\%)} 
        & 533~\textcolor{green!60!black}{(-28.3\%)} 
        & 3709~\textcolor{red!70!black}{(+23.6\%)} 
        & 154~\textcolor{red!70!black}{(+126.5\%)} \\
        \hspace{1.6em}xRAG
        & 1295~\textcolor{green!60!black}{(-7.2\%)} 
        & 1317~\textcolor{green!60!black}{(-56.1\%)} 
        & 70~\textcolor{red!70!black}{(+1.4\%)} 
        & 1309~\textcolor{red!70!black}{(+39.9\%)} 
        & 1190~\textcolor{green!60!black}{(-61.6\%)} 
        & 38~\textcolor{green!60!black}{(-47.9\%)} 
        & 1349~\textcolor{green!60!black}{(-5.7\%)} 
        & 1336~\textcolor{green!60!black}{(-53.6\%)} 
        & 74~\textcolor{red!70!black}{(+7.2\%)} 
        & 1208~\textcolor{red!70!black}{(+37.1\%)} 
        & 942~\textcolor{green!60!black}{(-65.0\%)} 
        & 78~\textcolor{red!70!black}{(+18.2\%)} 
        & 1188~\textcolor{red!70!black}{(+16.0\%)} 
        & 1158~\textcolor{green!60!black}{(-54.9\%)} 
        & 70~\textcolor{red!70!black}{(+11.1\%)} 
        & 844~\textcolor{red!70!black}{(+13.6\%)} 
        & 1336~\textcolor{green!60!black}{(-55.5\%)} 
        & 68~\textcolor{green!60!black}{(0.0\%)} \\
        \hspace{1.6em}COCOM
        & 488~\textcolor{green!60!black}{(-65.0\%)} 
        & 2346~\textcolor{green!60!black}{(-21.8\%)} 
        & 138~\textcolor{red!70!black}{(+100.0\%)} 
        & 368~\textcolor{green!60!black}{(-60.7\%)} 
        & 2459~\textcolor{green!60!black}{(-20.7\%)} 
        & 134~\textcolor{red!70!black}{(+83.6\%)} 
        & 642~\textcolor{green!60!black}{(-55.1\%)} 
        & 2311~\textcolor{green!60!black}{(-19.7\%)} 
        & 137~\textcolor{red!70!black}{(+98.6\%)} 
        & 514~\textcolor{green!60!black}{(-41.7\%)} 
        & 1841~\textcolor{green!60!black}{(-31.5\%)} 
        & 135~\textcolor{red!70!black}{(+104.5\%)} 
        & 540~\textcolor{green!60!black}{(-47.3\%)} 
        & 1905~\textcolor{green!60!black}{(-25.9\%)} 
        & 137~\textcolor{red!70!black}{(+117.5\%)} 
        & 478~\textcolor{green!60!black}{(-35.7\%)} 
        & 2335~\textcolor{green!60!black}{(-22.2\%)} 
        & 144~\textcolor{red!70!black}{(+111.8\%)} \\
        \hspace{1.6em}PISCO
        & 502~\textcolor{green!60!black}{(-64.0\%)} 
        & 2251~\textcolor{green!60!black}{(-25.0\%)} 
        & 142~\textcolor{red!70!black}{(+105.8\%)} 
        & 384~\textcolor{green!60!black}{(-59.0\%)} 
        & 2326~\textcolor{green!60!black}{(-25.0\%)} 
        & 139~\textcolor{red!70!black}{(+90.4\%)} 
        & 658~\textcolor{green!60!black}{(-54.0\%)} 
        & 2159~\textcolor{green!60!black}{(-25.0\%)} 
        & 140~\textcolor{red!70!black}{(+102.9\%)} 
        & 529~\textcolor{green!60!black}{(-40.0\%)} 
        & 2016~\textcolor{green!60!black}{(-25.0\%)} 
        & 136~\textcolor{red!70!black}{(+106.1\%)} 
        & 553~\textcolor{green!60!black}{(-46.0\%)} 
        & 1927~\textcolor{green!60!black}{(-25.0\%)} 
        & 141~\textcolor{red!70!black}{(+123.8\%)} 
        & 490~\textcolor{green!60!black}{(-34.1\%)} 
        & 2252~\textcolor{green!60!black}{(-25.0\%)} 
        & 145~\textcolor{red!70!black}{(+113.2\%)} \\
        \hspace{1.6em}\textbf{\system{}}
        & 535~\textcolor{green!60!black}{(-61.7\%)} 
        & 1166~\textcolor{green!60!black}{(-61.1\%)} 
        & 49~\textcolor{green!60!black}{(-29.0\%)} 
        & 528~\textcolor{green!60!black}{(-43.6\%)} 
        & 1270~\textcolor{green!60!black}{(-59.0\%)} 
        & 50~\textcolor{green!60!black}{(-31.5\%)} 
        & 565~\textcolor{green!60!black}{(-60.5\%)} 
        & 1144~\textcolor{green!60!black}{(-60.3\%)} 
        & 50~\textcolor{green!60!black}{(-27.5\%)} 
        & 504~\textcolor{green!60!black}{(-42.8\%)} 
        & 1120~\textcolor{green!60!black}{(-58.3\%)} 
        & 48~\textcolor{green!60!black}{(-27.3\%)} 
        & 496~\textcolor{green!60!black}{(-51.6\%)} 
        & 1271~\textcolor{green!60!black}{(-50.5\%)} 
        & 49~\textcolor{green!60!black}{(-22.2\%)} 
        & 434~\textcolor{green!60!black}{(-41.6\%)} 
        & 1384~\textcolor{green!60!black}{(-53.9\%)} 
        & 51~\textcolor{green!60!black}{(-25.0\%)} \\
        \multicolumn{19}{l}{\hspace{0.8em}\textit{\textbf{Top k=5}}} \\
        \hspace{1.6em}LLM (with RAG*) & 1537 & 10811 & 165 & 1498 & 11196 & 171 & 1514 & 10385 & 159 & 1458 & 9221 & 142 & 1459 & 4038 & 80 & 1533 & 10674 & 163 \\
        \hspace{1.6em}\textbf{\system{}} 
        & 590~\textcolor{green!60!black}{(-61.6\%)} 
        & 1664~\textcolor{green!60!black}{(-84.6\%)} 
        & 72~\textcolor{green!60!black}{(-56.4\%)} 
        & 534~\textcolor{green!60!black}{(-64.3\%)} 
        & 1764~\textcolor{green!60!black}{(-84.2\%)} 
        & 75~\textcolor{green!60!black}{(-56.1\%)} 
        & 1207~\textcolor{green!60!black}{(-20.3\%)} 
        & 1787~\textcolor{green!60!black}{(-82.8\%)} 
        & 73~\textcolor{green!60!black}{(-54.1\%)} 
        & 676~\textcolor{green!60!black}{(-53.6\%)} 
        & 1651~\textcolor{green!60!black}{(-82.1\%)} 
        & 76~\textcolor{green!60!black}{(-46.5\%)} 
        & 505~\textcolor{green!60!black}{(-65.4\%)} 
        & 1416~\textcolor{green!60!black}{(-64.9\%)} 
        & 53~\textcolor{green!60!black}{(-33.8\%)} 
        & 1050~\textcolor{green!60!black}{(-31.5\%)} 
        & 2034~\textcolor{green!60!black}{(-80.9\%)} 
        & 78~\textcolor{green!60!black}{(-52.1\%)} \\
        \midrule[0.8pt]

        \multicolumn{19}{l}{\textbf{Qwen2.5-7B-Instruct}} \\
        \multicolumn{19}{l}{\hspace{0.8em}\textit{\textbf{Top k=1}}} \\
        \hspace{1.6em}LLM (w/o RAG) & 792 & 983 & 36 & 386 & 918 & 35 & 851 & 990 & 35 & 551 & 852 & 37 & 449 & 950 & 36 & 627 & 933 & 36 \\
        \hspace{1.6em}LLM (with RAG) & 826 & 2749 & 44 & 368 & 2782 & 40 & 667 & 2785 & 38 & 661 & 2144 & 67 & 499 & 2273 & 40 & 524 & 2689 & 48 \\
        \hspace{1.6em}LLM (with RAG*) & 645 & 2576 & 64 & 426 & 2703 & 54 & 432 & 2526 & 51 & 622 & 2192 & 64 & 670 & 2220 & 65 & 491 & 2694 & 62 \\
        \hspace{1.6em}\textbf{\system{}} 
        & 433~\textcolor{green!60!black}{(-32.9\%)} 
        & 961~\textcolor{green!60!black}{(-62.7\%)} 
        & 48~\textcolor{green!60!black}{(-25.0\%)} 
        & 423~\textcolor{green!60!black}{(-0.7\%)} 
        & 1047~\textcolor{green!60!black}{(-61.3\%)} 
        & 48~\textcolor{green!60!black}{(-11.1\%)} 
        & 404~\textcolor{green!60!black}{(-6.5\%)} 
        & 947~\textcolor{green!60!black}{(-62.5\%)} 
        & 47~\textcolor{green!60!black}{(-7.8\%)} 
        & 490~\textcolor{green!60!black}{(-21.2\%)} 
        & 962~\textcolor{green!60!black}{(-56.1\%)} 
        & 58~\textcolor{green!60!black}{(-9.4\%)} 
        & 513~\textcolor{green!60!black}{(-23.4\%)} 
        & 1063~\textcolor{green!60!black}{(-52.1\%)} 
        & 53~\textcolor{green!60!black}{(-18.5\%)} 
        & 401~\textcolor{green!60!black}{(-18.3\%)} 
        & 1521~\textcolor{green!60!black}{(-43.5\%)} 
        & 50~\textcolor{green!60!black}{(-19.4\%)} \\
        \multicolumn{19}{l}{\hspace{0.8em}\textit{\textbf{Top k=5}}} \\
        \hspace{1.6em}LLM (with RAG*) & 733 & 9682 & 142 & 731 & 10004 & 147 & 734 & 9450 & 141 & 708 & 7989 & 121 & 691 & 3564 & 73 & 723 & 9522 & 151 \\
        \hspace{1.6em}\textbf{\system{}} 
        & 518~\textcolor{green!60!black}{(-29.3\%)} 
        & 1454~\textcolor{green!60!black}{(-85.0\%)} 
        & 65~\textcolor{green!60!black}{(-54.2\%)} 
        & 519~\textcolor{green!60!black}{(-29.0\%)} 
        & 1541~\textcolor{green!60!black}{(-84.6\%)} 
        & 66~\textcolor{green!60!black}{(-55.1\%)} 
        & 545~\textcolor{green!60!black}{(-25.7\%)} 
        & 1401~\textcolor{green!60!black}{(-85.2\%)} 
        & 69~\textcolor{green!60!black}{(-51.1\%)} 
        & 477~\textcolor{green!60!black}{(-32.6\%)} 
        & 1478~\textcolor{green!60!black}{(-81.5\%)} 
        & 73~\textcolor{green!60!black}{(-39.7\%)} 
        & 515~\textcolor{green!60!black}{(-25.5\%)} 
        & 1228~\textcolor{green!60!black}{(-65.5\%)} 
        & 58~\textcolor{green!60!black}{(-20.5\%)} 
        & 589~\textcolor{green!60!black}{(-18.5\%)} 
        & 1663~\textcolor{green!60!black}{(-82.5\%)} 
        & 71~\textcolor{green!60!black}{(-53.0\%)} \\
        \bottomrule
    \end{tabular}
    }
\end{table*}
\section{Case Study}\label{appendix_i}
Examples of full compression not following instructions:
\begin{tcolorbox}[colback=white, colframe=blue!50!black, boxsep=2pt, 
    left=2pt, right=2pt,   
    top=2pt, bottom=2pt,   
    before skip=2pt, after skip=2pt, 
    fontupper=\footnotesize, fonttitle=\footnotesize]
    \tiny
    \textbf{\underline{Input Prompt:}}
    
    Background: DOCUMENT
    
    Who is Republic of Hawaii's co-founder?\\[0.3\baselineskip]
    \textbf{\underline{Generator Output under full-compression:}}
    
    \textcolor{red}{Title: O. M. Hustvedt}
    
    \textcolor{red}{Content: ...}\\[0.3\baselineskip]
    \textbf{\underline{Generator Output without compression:}}
    
    \textcolor{commentgreen}{The Republic of Hawaii's co-founder mentioned in the background information is Henry E. Cooper.}
\end{tcolorbox}
\begin{tcolorbox}[colback=white, colframe=blue!50!black, boxsep=2pt, 
    left=2pt, right=2pt,   
    top=2pt, bottom=2pt,  
    before skip=2pt, after skip=2pt, 
    fontupper=\footnotesize, fonttitle=\footnotesize]
    \tiny
    \textbf{\underline{Input Prompt:}}
    
    Background: DOCUMENT
    
    Why does Olaf M. Hustvedt\'s son dropped from the academy?\\[0.3\baselineskip]
    \textbf{\underline{Generator Output under full-compression:}}
    
    \textcolor{red}{Title: Olaf M. Hustvedt}
    
    \textcolor{red}{Content: ...}\\[0.3\baselineskip]
    \textbf{\underline{Generator Output without compression:}}
    
    \textcolor{commentgreen}{According to the information provided, Olaf M. Hustvedt's son, Erling H. Hustvedt, was dropped from the U.S. Naval Academy due to an incorrect evaluation of his eyesight.}
\end{tcolorbox}
Cases of redundant/misleading retrieved documents:
\begin{tcolorbox}[colback=white, colframe=blue!50!black, boxsep=2pt, 
    left=2pt, right=2pt,   
    top=2pt, bottom=2pt,  
    before skip=2pt, after skip=2pt,
    fontupper=\footnotesize, fonttitle=\footnotesize]
    \tiny
    \textbf{\underline{Document:}}

    Title: God of War (2018 video game)
    
    Content: God of War is ... \textcolor{commentgreen}{It was released worldwide on April 20, 2018}, for the PlayStation 4 (PS4) with a Microsoft Windows version set to release on January 14, 2022. The game is the eighth installment in the God of War series, the eighth chronologically, and the sequel to \textcolor{red}{2010's God of War III}.\\[0.3\baselineskip]
    \textbf{\underline{Input Prompt:}}
    when was the last god of war made?\\[0.3\baselineskip]
    \textbf{\underline{Baseline RAG Output:}}
    \textcolor{red}{2010}\\[0.3\baselineskip]
    \textbf{\underline{\system{} Output:}}
    \textcolor{commentgreen}{2018}
\end{tcolorbox}
\begin{tcolorbox}[colback=white, colframe=blue!50!black, boxsep=2pt, 
    left=2pt, right=2pt,  
    top=2pt, bottom=2pt, 
    before skip=2pt, after skip=2pt,
    fontupper=\footnotesize, fonttitle=\footnotesize]
    \tiny
    \textbf{\underline{Document:}}

    Title: The A-Team
    
    Content:  Although the part of Face was written by Frank Lupo and Stephen J. Cannell \textcolor{commentgreen}{with Dirk Benedict in mind}, NBC insisted that the part should be played by another actor. Therefore, in the pilot, \textcolor{red}{Face was portrayed by Tim Dunigan}, \textcolor{commentgreen}{who was later replaced by Benedict}, with the comment that \textcolor{red}{Dunigan was "too tall and too young"}.\\[0.3\baselineskip]
    \textbf{\underline{Input Prompt:}}
    who played the face in the a team?\\[0.3\baselineskip]
    \textbf{\underline{Baseline RAG Output:}}
    \textcolor{red}{Tim Dunigan}\\[0.3\baselineskip]
    \textbf{\underline{\system{} Output:}}
    \textcolor{commentgreen}{Dirk Benedict}
\end{tcolorbox}
\section{Prompts}\label{appendix_j}
\begin{tcolorbox}[colback=white, colframe=blue!50!black, boxsep=1pt, 
    left=1pt, right=1pt,  
    top=1pt, bottom=1pt,  
    before skip=1pt, after skip=1pt,
    fontupper=\footnotesize, fonttitle=\footnotesize]
    \tiny
    \textbf{$P_{select}$:}

    \#\#\# Document

    [Document]

    \#\#\# Question

    [Question]
    
    \#\#\# Instruction

    Extract the key information from the document that is helpful to answer the question.

    \textcolor{blue!70!black}{<ENCODE>}...\textcolor{blue!70!black}{<ENCODE>}
\end{tcolorbox}
\begin{tcolorbox}[colback=white, colframe=blue!50!black, boxsep=1pt, 
    left=1pt, right=1pt,  
    top=1pt, bottom=1pt,   
    before skip=1pt, after skip=1pt, 
    fontupper=\footnotesize, fonttitle=\footnotesize]
    \tiny
    \textbf{$P_{gen}$:}
    
    \#\#\# Reference

    \textcolor{green!70!black}{<DOCUMENT\_START>}\textcolor{red!70!black}{<DOCUMENT>}...\textcolor{red!70!black}{<DOCUMENT>}\textcolor{green!70!black}{<DOCUMENT\_END>}

    \#\#\# Question

    [Question]

    \#\#\# Instruction

    Answer the question according to the reference provided above.

    \#\#\# Restriction

    1. You must use English.

    2. You must DIRECTLY provide the answer in this STRICT format: <answer></answer>.

    3. You must not generate any other text.
\end{tcolorbox}
\begin{tcolorbox}[colback=white, colframe=blue!50!black, boxsep=1pt, 
    left=1pt, right=1pt,  
    top=1pt, bottom=1pt, 
    before skip=1pt, after skip=1pt, 
    fontupper=\footnotesize, fonttitle=\footnotesize]
    \tiny
    \textbf{Prompt for document content filtering}:
    
    \#\#\# Instruction

    You will be given a document. Verify if the document is a piece of human-readable plain text.

    \#\#\# Document

    [DOCUMENT]

    \#\#\# Criteria

    ALL PART of the document must look like a piece of descriptive text. If it seems to be a piece of code, table, figure or other non-descriptive elements, it is considered of BAD quality. You must be a strict judge and consider a text as BAD at any opportunity.

    \#\#\# Restriction

    1. You must generate your judgement in this STRICT format: <document></document>

    2. You must not generate any other text.
\end{tcolorbox}
\begin{tcolorbox}[colback=white, colframe=blue!50!black, boxsep=1pt, 
    left=1pt, right=1pt,  
    top=1pt, bottom=1pt,   
    before skip=1pt, after skip=1pt, 
    fontupper=\footnotesize, fonttitle=\footnotesize]
    \tiny
    \textbf{Prompt for document quality filtering}:
    
    \#\#\# Instruction

    You will be given a document. Evaluate the amount of information contained by the document.

    \#\#\# Document

    [DOCUMENT]

    \#\#\# Criteria

    You will score the amount of information of the document as an integer in 1,2,3,4,5,6,7,8,9,10.

    \#\#\# Format restriction

    1. You must generate your evaluation in this STRICT format: <response></response>

    2. You must not generate any other text.
\end{tcolorbox}
\begin{tcolorbox}[colback=white, colframe=blue!50!black, boxsep=1pt, 
    left=1pt, right=1pt,   
    top=1pt, bottom=1pt,  
    before skip=1pt, after skip=1pt,
    fontupper=\footnotesize, fonttitle=\footnotesize]
    \tiny
    \textbf{Prompts for EASY QA generation}:
    
    \#\#\# Document

    [DOCUMENT]

    \#\#\# Instruction

    Ask a question about one single fact mentioned in the above document, and provide its answer. The answer should be short, concise and factual (1-10 words typically) and correctly answers the question.

    \#\#\# Restriction

    1. You must use English.

    2. You must generate the question and its answer in this STRICT format: <question></question><answer></answer>.

    3. You must not generate any other text.\\[0.3\baselineskip]
    \textbf{Prompts for HARD QA generation}:
    
    \#\#\# Document

    [DOCUMENT]

    \#\#\# Instruction

    Ask a challenging, real-world question about one single fact that can be inferred from the above document and provide its answer. The question should require reasoning, inference, or common sense beyond just looking up facts directly from the text. The question should never ask for more than one fact. The answer should be short, concise and factual (1-10 words typically) and correctly answers the question. Avoid questions that can be answered by simply copying text from the document.

    \#\#\# Restriction

    1. You must use English.

    2. You must generate the question and its answer in this STRICT format: <question></question><answer></answer>.

    3. You must not generate any other text.
\end{tcolorbox}
\begin{tcolorbox}[colback=white, colframe=blue!50!black, boxsep=1pt, 
    left=1pt, right=1pt, 
    top=1pt, bottom=1pt,   
    before skip=1pt, after skip=1pt, 
    fontupper=\footnotesize, fonttitle=\footnotesize]
    \tiny
    \textbf{Prompt for QA filtering:}
    
    \#\#\# Instruction

    You will be given a document, a question on this document, and an answer to this question. Judge the correctness of the answer, then evaluate the difficulty of the question to be correctly answered given the document content.

    \#\#\# Document

    [DOCUMENT]

    \#\#\# Question

    [Question]

    \#\#\# Answer

    [Answer]

    \#\#\# Criteria

    - Judge the answer by if it is correct to answer the question. 
    
    If correct, CORRECT. If wrong, WRONG.

    - Then score the difficulty of the question to be correctly answered given the document as an integer in 1,2,3,4,5.

    \#\#\# Format restriction

    1. You must generate your response in this STRICT format: <correctness></correctness><difficulty></difficulty>.

    2. You must not generate any other text.
\end{tcolorbox}
\begin{tcolorbox}[colback=white, colframe=blue!50!black, boxsep=1pt, 
    left=1pt, right=1pt,   
    top=1pt, bottom=1pt,  
    before skip=1pt, after skip=1pt,
    fontupper=\footnotesize, fonttitle=\footnotesize]
    \tiny
    \textbf{Prompt of LLM-as-a-judge metric:}
    
    You will be given a question, a candidate answer, and a reference answer. Judge whether the candidate answer correctly addresses the question compared to the reference. Respond **ONLY** with a numeric score between 0 (completely wrong) and 1 (perfectly correct).

    \#\#\# Question
    
    [Question]

    \#\#\# Candidate Answer
    
    [Prediction]

    \#\#\# Reference Answer
    
    [Ground truth]
\end{tcolorbox}

\end{document}